\documentclass[conference]{IEEEtran}
\IEEEoverridecommandlockouts

\usepackage{cite}
\usepackage{amsmath,amssymb,amsfonts}
\usepackage{algorithm}
\usepackage{algorithmic}
\usepackage{graphicx}
\usepackage{textcomp}
\usepackage{xcolor}
\usepackage{booktabs}
\usepackage{multirow}
\usepackage{subcaption}
\usepackage{makecell}
\usepackage{tabularx}
\usepackage{float}
\usepackage[table]{xcolor}

\newcommand{\name}{\textsc{Gen}}

\def\BibTeX{{\rm B\kern-.05em{\sc i\kern-.025em b}\kern-.08em
    T\kern-.1667em\lower.7ex\hbox{E}\kern-.125emX}}
\begin{document}

\title{Rethinking Flexible Graph Similarity Computation: One-step Alignment with Global Guidance}

\author{
\centering
\IEEEauthorblockN{Zhouyang Liu}
\IEEEauthorblockA{\textit{College of Computer Science and Technology} \\
\textit{National University of Defense Technology}\\
Hunan, China \\
liuzhouyang20@nudt.edu.cn}
\\

\IEEEauthorblockN{Yixin Chen}
\IEEEauthorblockA{\textit{College of Computer Science and Technology} \\
\textit{National University of Defense Technology}\\
Hunan, China \\
chenyixin@nudt.edu.cn}
\\

\IEEEauthorblockN{Shuai Ma}
\IEEEauthorblockA{
\textit{Beihang University}\\
Beijing, China \\
mashuai@buaa.edu.cn}
\and
\IEEEauthorblockN{Ning Liu}
\IEEEauthorblockA{
\textit{Information Support Force Engineering University}\\
Hubei, China \\
liuning17a@nudt.edu.cn}
\\

\IEEEauthorblockN{Jiezhong He}
\IEEEauthorblockA{\textit{College of Computer Science and Technology} \\
\textit{National University of Defense Technology}\\
Hunan, China \\
jiezhonghe@nudt.edu.cn}
\\

\IEEEauthorblockN{Dongsheng Li}
\IEEEauthorblockA{\textit{College of Computer Science and Technology} \\
\textit{National University of Defense Technology}\\
Hunan, China \\
dsli@nudt.edu.cn}
}

\maketitle

\begin{abstract}
Graph Edit Distance (GED) is a widely used measure of graph similarity, valued for its flexibility in encoding domain knowledge through operation costs. However, existing learning-based approximation methods follow a modeling paradigm that decouples local candidate match selection from both operation costs and global dependencies between matches. This decoupling undermines their ability to capture the intrinsic flexibility of GED and often forces them to rely on costly iterative refinement to obtain accurate alignments. In this work, we revisit the formulation of GED and revise the prevailing paradigm, and propose Graph Edit Network (\name), an implementation of the revised formulation that tightly integrates cost-aware expense estimation with globally guided one-step alignment. Specifically, \name\ incorporates operation costs into node matching expenses estimation, ensuring match decisions respect the specified cost setting. Furthermore, \name\ models match dependencies within and across graphs, capturing each match’s impact on the overall alignment. These designs enable accurate GED approximation without iterative refinement. Extensive experiments on real-world and synthetic benchmarks demonstrate that \name\ achieves up to a 37.8\% reduction in GED predictive errors, while increasing inference throughput by up to 414×. These results highlight \name's practical efficiency and the effectiveness of the revision. Beyond this implementation, our revision provides a principled framework for advancing learning-based GED approximation.
\end{abstract}

\section{Introduction}
Graph similarity computation is a fundamental problem in graph data management, supporting tasks such as similarity search, nearest neighbor retrieval, and query processing \cite{search, database2, survey, database3, database5, database, database4,nciso,g2r}. A widely used measure for this task is Graph Edit Distance (GED), which quantifies the minimum edit expense required to transform one graph into another through a sequence of atomic edit operations. In this context, the edit expense is the sum of operation costs over the sequence, and the atomic operations include node or edge insertion, deletion, or substitution. Among  similarity measures, GED is distinctive for its flexibility: operation costs can be customized to encode domain knowledge. For example, in molecular databases, higher substitution costs may be assigned to core atoms than to peripheral ones, ensuring that top-ranked retrievals preserve essential structural patterns.

\begin{figure}
    \includegraphics[width=\linewidth]{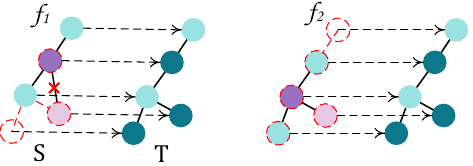}
    \caption{Variations in operation costs lead to different optimal alignments. The black dashed lines are matches, and the red markings are edits. \( f_1 \) and \( f_2\) are mappings. 
    \( f_1 \) incurs 2 node substitutions (ns), 1 node insertion, 1 edge deletion, and 2 edge insertions (ei), denoted as \([2, 1, 0, 1, 2]\), while \( f_2 \) incurs \([4, 1, 0, 0, 1]\). When ns costs 2, other operations cost 1, \( f_1 \) is optimal with a expense of 8, while \( f_2 \) costs 10. However, if ei costs 2, others cost 1, \( f_1 \) costs 8 while \( f_2 \) costs 7, making \( f_2 \) the optimal choice.
    } \label{fig:motive}
\end{figure}

Classical algorithms, such as A* search \cite{A*GED, search, ged1, a*lsa}, explore possible node matches to find alignments with minimum edit expense under a given cost setting. 
Variations in cost settings can therefore lead to different optimal alignments, reflecting GED’s flexibility, as illustrated in Figure \ref{fig:motive}. However, due to its NP-hard complexity \cite{nphard}, exact GED computation is feasible only for small graph databases. Recent studies show that even the most advanced exact algorithms fail to finish GED computation within a reasonable time for graphs with as few as 16 nodes \cite{exact}. Moreover, these algorithms operate in an on-the-fly manner, preventing the reuse of intermediate results and hindering index construction. These limitations make them impractical for modern applications involving high-volume graph collections and high-frequency querying.

To overcome these limitations, recent research has explored latent space GED approximation \cite{tagsim, gmn, genn, egsc, greed, eric, simgnn, graphsim, graphedx, gedgnn, prune4sed, h2mn, GED-CDA, noah, mata*}. They first obtain node embeddings through techniques such as Graph Neural Networks (GNNs) \cite{GIN,gat,gatv2,gcn}. Based on these embeddings, existing approaches fall into two categories: (1) \textbf{Coarse-grained graph-level approaches}, which compress node representations into fixed-length vectors. While efficient, compressing node information inevitably loses node-level structural details. 
This absence of information reduces the ability to model potential edit operations, limiting the capacity to capture GED’s flexibility. (2) \textbf{Fine-grained node-level methods}, which approximate matching expenses for candidate matches by computing cross-graph node embedding distances. During the alignment process, these expenses are used to guide iterative refinement procedures, typically through differentiable variants of the Hungarian or Sinkhorn algorithms. By assigning higher weights to low-expense matches, they progressively minimize the overall edit expense. 

Although fine-grained methods generally have quadratic computational complexity, node-level details give them greater potential to capture GED’s flexibility. To exploit this potential, it is necessary to understand how varying cost settings influence candidate match selection and how each candidate match contributes to the alignment. Since the overall expense depends on the expenses of individual selected matches, finding an optimal alignment can be broken down into candidate match selection problems. We therefore revisit GED from a matching perspective and identify three key interacting factors: (i) intrinsic differences between nodes, including attributes and neighborhood structures; (ii) the type and cost of operations needed to resolve these differences; and (iii) dependencies between candidate matches and other established matches.

Rethinking existing fine-grained learning-based methods through this lens reveals two issues in the prevailing paradigm. First, matching expenses are typically computed from embedding differences, while the specified cost setting is applied only after alignments are formed. For example, GraphEdX \cite{graphedx} incorporates operation costs only in the final GED computation. This decoupling breaks the link between candidate match selection and cost settings, yielding suboptimal alignments irrelevant to costs. 
Second, these methods rely solely on local embedding distances to guide matching, ignoring dependencies between candidate matches. As a result, matches are often selected in isolation and require iterative corrections, which improve accuracy but also increase computation and latency. Collectively, these issues indicate a structural limitation that the prevailing paradigm decouples operation costs, intrinsic differences, and match dependencies during candidate match selection, motivating a principled revision that integrates these factors within a single framework to reconcile efficiency with fine-grained edit operation modeling.

In this work, we revisit the formulation of GED and rethink the prevailing paradigm of fine-grained approximation. Unlike prior studies that focus primarily on on embedding, scoring, and alignment designs, we systematically dissect the factors determining candidate match selection and reveal a structural limitation in existing methods: the decoupling of operation costs, intrinsic differences, and dependencies between matches. Based on this insight, we propose a revised formulation and introduce the Graph Edit Network (\name) as an implementation. \name\ consists of two complementary modules: (1) Cost-aware matching expense estimation, which reweigh intrinsic node embedding differences using operation costs, 
coupling operation costs and intrinsic differences with node matching expenses. 
(2) One-step node alignment with global guidance, which evaluate each candidate match’s impact on the alignment, considering its expense and dependencies with other matches within and between graphs, 
coupling matching expenses and match dependencies with matching decisions.
The main contributions of this study are:
\begin{itemize}
\item We rethink the prevailing paradigm for GED approximation, identify its structural limitation, and propose a principled revision that integrates operation costs, intrinsic differences, and match dependencies within a single framework.
\item We incorporate operation costs into node matching expenses by reweighing intrinsic node embedding differences, ensuring that matching decisions respect the specified cost setting. 
\item We construct global guidance by estimating each candidate match’s impact on alignment, considering its expense and dependencies with other matches, enabling one-step alignment without iterative refinement. 
\item Extensive experiments show that \name\ outperforms state-of-the-art methods in accuracy and efficiency, reducing predictive error by up to 37.8\% and increasing inference throughput by up to 414×, demonstrating strong potential for high-volume and high-frequency querying scenarios.
\end{itemize}

\section{Related Work}
Learning-based graph similarity computation (GSC) aims to approximate similarity measures such as graph edit distance (GED) using neural networks~\cite{simgnn, gmn}. We focus on end-to-end GED approximation methods, which can be divided into coarse-grained graph-level and fine-grained node-level approaches. For details, we refer the reader to recent surveys~\cite{survey}.

\paragraph{Coarse-grained graph-level methods} These methods embed entire graphs into latent spaces and compute similarity based on graph representations~\cite{gmn,egsc,eric,greed,tagsim}. Their primary focus lies in enhancing the expressiveness of graph representations while maintaining efficiency. For example, GMN-emb~\cite{gmn} employs a gated mechanism to highlight important nodes during embedding. To balance prediction accuracy and storage efficiency, EGSC~\cite{egsc} computes attention-based pair-dependent representations and further introduces a teacher-student framework to learn pair-independent ones. Eric~\cite{eric} enforces node-graph alignment during training but detaches it during inference, thereby decoupling alignment from prediction to improve efficiency. Greed~\cite{greed} incorporates the inductive bias of GED by modeling the triangle inequality among graph embeddings, which accelerates convergence and improves performance. TaGSim~\cite{tagsim} enhances interpretability by predicting graph edit vectors (GEVs) that explicitly represent edit operation counts. However, despite these advancements, graph-level methods inevitably lose node-level structural details that are critical for modeling edit operations, and thus fail to capture the intrinsic flexibility of GED.

\paragraph{Node-level comparison methods} These methods compare node embeddings across graphs to capture fine-grained discrepancies, and typically estimate GED either from the distribution of pairwise distances or by constructing alignments that minimize total cross-graph node distance~\cite{gmn, simgnn, graphsim, gotsim, gedgnn, graphedx,sigmod1,vldb1}. Early studies aim to address the lack of match dependencies caused by node permutation invariance. SimGNN~\cite{simgnn} generates histogram features from node-pair similarities to capture distributional information, partially compensating this lack. Although the model is end-to-end trainable, these histogram features are non-differentiable and cannot participate in gradient-based optimization. GraphSim~\cite{graphsim} reorders multi-scale similarity matrices via BFS traversal to impose a canonical node order, but the results remain sensitive to the ordering.

Later works move toward explicit alignment. GMN-match~\cite{gmn} introduces an iterative fuse-refine mechanism: it fuses cross-graph node embeddings based on distances and refines them using neighborhood representations. Following this idea, subsequent methods perform alignment optimization in matrix form under combinatorial constraints. GOTSim~\cite{gotsim} formulates GED as an optimal transport problem, augments the distance matrix to handle insertions and deletions, and solves alignment using a CPU-based linear assignment solver. To overcome the efficiency bottleneck of CPU-based solvers, GEDGNN~\cite{gedgnn}, GraphEdX~\cite{graphedx}, MATA*~\cite{mata*}, GEDIOT\cite{sigmod1} and FGWAlign\cite{vldb1} employ Gumbel-Sinkhorn networks for iterative refinement of soft alignments. Among them, GEDGNN and MATA* leverage pretrained models to guide exact GED computation, GraphEdX defines multiple operation-specific distance measures and incorporates operation costs to reweigh alignments, and GEDIOT and FGWAlign introduce Gromov-Wasserstein distance. Despite these variations, all these methods follow the same paradigm: derive node distances from embeddings and iteratively refine alignments using them as local guidance, leaving operation costs and match dependencies largely unmodeled. These structural limitations motivate a principled revision that integrates cost-aware alignment with global guidance for more accurate GED approximation.

\begin{table}[t]
\centering
\caption{Summary of notations used in this paper.}
\label{tab:notation}
\resizebox{\linewidth}{!}{ \renewcommand{\arraystretch}{1.0}
\footnotesize 
\begin{tabularx}{\linewidth}{lX}
\toprule
\textbf{Notation} & \textbf{Description} \\
\midrule
$\mathcal{Q}, \mathcal{D}$ & Query and data graph collections \\
$G=(\mathcal{V}_G, \mathcal{E}_G)$ & a graph with node sets $\mathcal{V}$ and edge sets $\mathcal{E}$ \\
$\mathbf{A}_G$ & Adjacency matrix of $G$ \\
$\text{GED}(Q,D)$ & Graph Edit Distance between $Q$ and $D$  \\
$dist_{q,d}$ & intrinsic difference between elements $q \in \mathcal{V}_Q \cup \mathcal{E}_Q$ and $d \in \mathcal{V}_D \cup \mathcal{E}_D$ \\
$c_{q,d}$ & Matching expense between node $q$ and $d$, combining operation cost and intrinsic difference \\
$\text{cost}_{\text{op}_{q,d}}$ & Numeric cost of applying atomic operation $\text{op}$ on node/edge $q \to d$ \\
$f$ & A valid sequence of atomic edit operations or node alignment between $Q$ and $D$ \\
$\mathcal{F}$ & Set of all valid edit sequences (or alignments) \\
$\epsilon$ & Null element representing a virtual node or missing edge \\
$\Phi(\cdot)$ & Node encoder (typically a GNN) mapping nodes to embedding space \\
$\Psi(\cdot,\cdot)$ & Pairwise distance function between node embeddings \\
$c_{\mathbf{x}_q, \mathbf{x}_d}$ & Node-level matching expense estimated from embeddings of $q$ and $d$ \\
$w_{q,d}^{(k)}$ & Weight of candidate match $(q,d)$ at iteration $k$ \\
$\mathbf{c}(Q,D)$ & Total estimated matching expense used to approximate $\text{GED}(Q,D)$ \\
\bottomrule
\end{tabularx}}
\end{table}

\section{Preliminaries}
\paragraph{Problem Formulation} 
Given a query graph collection $\mathcal{Q}$ and a data graph collection $\mathcal{D}$, the goal of flexible graph similarity computation is to evaluate the graph edit distance $\text{GED}(Q, D) \in [0, +\infty)$ for all $Q \in \mathcal{Q}$ and $D \in \mathcal{D}$ under a specified operation cost setting. Here, $Q = (\mathcal{V}_Q, \mathcal{E}_Q)$ and $D = (\mathcal{V}_D, \mathcal{E}_D)$, where $\mathcal{V}_Q$ and $\mathcal{V}_D$ denote node sets and $\mathcal{E}_Q$ and $\mathcal{E}_D$ denote edge sets, which can equivalently be represented by adjacency matrices $\mathbf{A}_Q$ and $\mathbf{A}_D$. The cost setting assigns a numeric cost $\text{cost}_{\text{op}_{q,d}}$ to each atomic operation $\text{op}$ that modifies an element $q \in \mathcal{V}_Q \cup \mathcal{E}_Q$ to $d \in \mathcal{V}_D \cup \mathcal{E}_D$. In this work, we focus on undirected graphs without edge labels, where the permitted atomic edit operations include node insertion, deletion, and substitution, as well as edge insertion and deletion.

\paragraph{Graph Edit Distance (GED)}
Given a query graph $Q$ and a data graph $D$,  if the two graphs have different numbers of nodes, virtual nodes can be added to the smaller graph so that $|\mathcal{V}_Q| = |\mathcal{V}_D|$. The graph edit distance $\text{GED}(Q,D)$ is defined as the minimum total cost of transforming $Q$ into $D$:
\begin{align}
&c_{q,d} = \text{cost}_{\text{op}_{q,d}} \cdot dist_{q,d},\label{eq:gen_cost} \\
&\textbf{c}(Q,D)_f = \sum_{(q,d) \in f} c_{q,d}, \label{eq:gen_total_cost1} \\
&\text{GED}(Q,D) = \min_{f \in \mathcal{F}} \textbf{c}(Q,D)_f,
\end{align}
where $dist_{q,d}$ quantifies the intrinsic difference between $q$ and $d$ (e.g., label mismatch or edge inconsistency), $c_{q,d}$ is the matching expense, $f$ denotes a valid sequence of atomic operations, and $\mathcal{F}$ is the set of all such sequences. Under a uniform cost setting, where all atomic operations share the same cost, $\text{GED}(Q,D)$ is symmetric, i.e., $\text{GED}(Q,D) = \text{GED}(D,Q)$. In contrast, a non-uniform cost setting generally leads to $\text{GED}(Q,D) \neq \text{GED}(D,Q)$.

Since a valid edit path can be equivalently represented by a node alignment, Eq.~\eqref{eq:gen_total_cost1} can be reformulated as
\begin{align}
\textbf{c}(Q,D)_f = \sum_{(q,d) \in f} c_{q,d} 
+ \sum_{e \in f(\mathbf{A}_Q) \setminus \mathbf{A}_D} c_{e,\epsilon} 
+ \sum_{e' \in \mathbf{A}_D \setminus f(\mathbf{A}_Q)} c_{\epsilon,e'}, \label{eq:gen_total_cost2}
\end{align}
where $f$ is the alignment derived from a valid edit path, $\mathbf{A}_Q$ and $\mathbf{A}_D$ represent intra-graph dependencies, and $\epsilon$ denotes the absence of a edge. The second and third terms correspond to inter-graph discrepancies, i.e., edges deleted or inserted under alignment $f$. In summary, the matching of each candidate pair is influenced jointly by intrinsic differences, operation costs, and structural dependencies, reflecting how local node matches impact on the global graph alignment.

\section{Rethinking Flexible GED Computation in Embedding Space}\label{sec:gen_analysis}

Fine-grained node-level methods for GED approximation generally follow three stages:
\textbf{(1) Encoding.} An injective encoder $\Phi: \mathbf{X} \mapsto \mathbb{E}^{|\mathcal{V}|\times H}$ maps each node’s local structure into a $H$-dimensional embedding space, where $|\mathcal{V}|$ is the number of nodes.
\textbf{(2) Scoring.} A distance function $\Psi(\cdot,\cdot)$ computes pairwise distances between embeddings of query and data graph nodes, forming a distance matrix.
\textbf{(3) Alignment.} To simplify matching, feasibility constraints are relaxed by discarding strict one-to-one correspondence. Embedding distances are then used to guide the weight computation of candidate matches, and the weighted sum of distances serves as an approximation of the GED.
\begin{align}
&c_{\mathbf{x}_q,\mathbf{x}_d} = \text{Estimator}\Big(\Psi\big(\Phi(q), \Phi(d)\big)\Big),\label{eq:gen_gnn_cost}\\
&w_{q,d}^{(k+1)} = \text{WeightUpdate}\Big(w^{(k)}_{q,d}, c_{\mathbf{x}_q,\mathbf{x}_d}\Big),\label{eq:gen_weight} \\
&\textbf{c}(Q,D) = \text{cost}_{q,d} \cdot \sum_{q,d} w_{q,d}^K \cdot c_{\mathbf{x}_q,\mathbf{x}_d}.\label{eq:gen_total_cost3}
\end{align}
Here, $\Phi(\cdot)$ is usually a stacked Graph Neural Network (GNN) that embeds each node using its local structural context, and $\text{Estimator}(\cdot)$ is typically a simple MLP or identity mapping. The resulting $c_{\mathbf{x}_q,\mathbf{x}_d}\in\mathbb{R}^H$ is the estimated node-level matching expense between $q$ and $d$. The initial weight $w^{(0)}$ for each candidate match is set to one. At each iteration $k+1$, the weight $w_{q,d}^{(k+1)}$ is updated via the $\text{WeightUpdate}$ function, depending on the previous weight matrix and the guidance $c_{\mathbf{x}_q,\mathbf{x}_d}$. After $K$ iterations, the total matching expense is computed as the weighted sum of node-level matching expenses, scaled by the operation cost vector $\text{cost}_{q,d}\in\mathbb{R}^{|\text{op}|}$, which records the cost of all atomic operations between $q$ and $d$. This yields the estimated matching expense $\textbf{c}(Q,D)$, which can be directly used to approximate $\text{GED}(Q,D)$ without combinatorial search. In the following, we write $c_{\mathbf{x}_q,\mathbf{x}_d}$ as $c_{q,d}$ for simplicity.

 Let $GED = \langle \Pi^*, C^* \rangle$ be the exact distance and $\widehat{GED} = \langle W, C \rangle$ be the neural approximation. Following the analysis in \cite{sigmod1}, the total error is bounded by:
\begin{equation}
\label{eq:bound}
|\widehat{GED} - GED| \le \underbrace{\|W\|_F \cdot \|\Delta C\|_F}_{\text{Term 1: Expense Gap}} + \underbrace{\|\Delta W\|_F \cdot \|C^*\|_F}_{\text{Term 2: Alignment Gap}}
\end{equation}
where $\Delta C = C^* - C$ and $\Delta W = \Pi^* - W$ represent the discrepancies in matching expense estimation and structural alignment, respectively. Ideally, a tight approximation requires the concurrent minimization of these two terms.

However, comparing Equations~(\ref{eq:gen_gnn_cost}) and (\ref{eq:gen_total_cost3}) with Equation~(\ref{eq:gen_cost}) shows that existing fine-grained methods follow a modeling paradigm that fundamentally decouples node alignment from operation costs. Specifically, $\text{cost}_{q,d}$ is applied only after node alignment, while $c_{q,d}$ reflects only intrinsic node differences. As a result, alignments neglect operation costs, leading to an inaccurate expense matrix $C$ and yielding a significant \textbf{Expense Gap}. Moreover, the comparison between Equation~(\ref{eq:gen_total_cost2}) and Equation (\ref{eq:gen_weight}) shows that candidate selection is also decoupled from match dependencies: matching weights rely solely on local node distances, insufficiently encoding intra-graph and inter-graph dependencies, which can lead to locally optimal but globally suboptimal matches, thereby increasing the \textbf{Alignment Gap}. To compensate, existing methods often require iterative refinement, increasing computational cost. Collectively, these issues reveal a fundamental limitation of the prevailing GED approximation paradigm.
\begin{figure}
    \centering
    \includegraphics[width=0.8\linewidth]{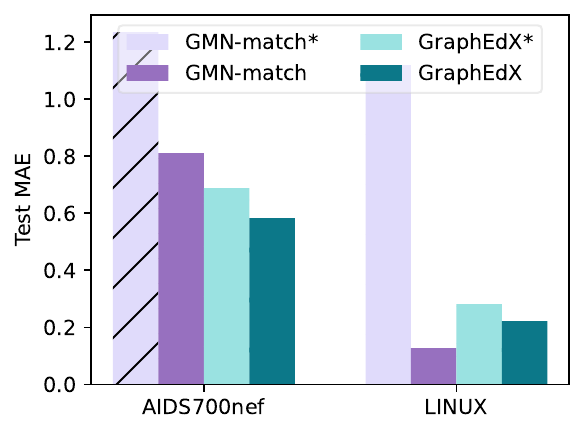}
    \caption{Models marked with ‘*’ perform only a single round of refinement.
    } \label{fig:refine}
\end{figure}

To empirically illustrate the dependence on iterative refinements, we conducted a study shown in Figure~\ref{fig:refine}. Limiting these models to a single iteration (marked with “*”) increases MAE by 15\% to 89\% compared with the authors’ recommended default settings, directly demonstrating the critical role of iterative refinement in mitigating the paradigm’s intrinsic limitations. By revisiting flexible GED computation, we argue that effective fine-grained GED approximation should follow a paradigm that aligned with GED principles. Candidate match selection should consider three interacting factors: intrinsic node differences, operation costs, and global matching dependencies. Incorporating these factors could provide a more comprehensive global guidance signal, potentially reducing the reliance on costly iterative refinements and improving both alignment quality and computational efficiency.

\begin{figure*}[!ht]
    \centering
    \includegraphics[width=\linewidth]{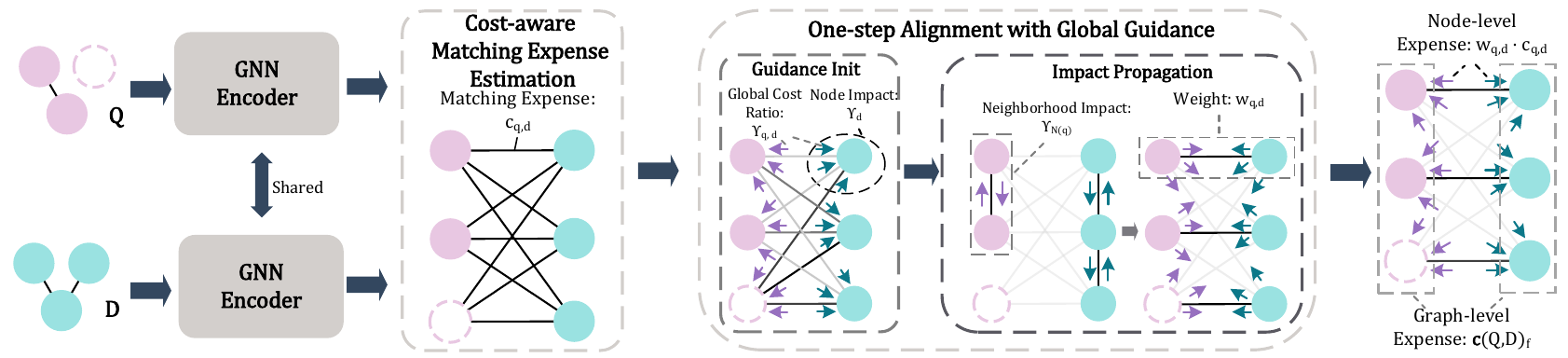}
    \caption{Overview of \name. The dashed node represents the dummy node.} 
    \label{fig:overview}
\end{figure*}
\section{Methodology}

To account for intrinsic node differences, operation costs, and global matching dependencies in candidate match selection, we propose a principled modification of Equations~(\ref{eq:gen_gnn_cost}), (\ref{eq:gen_weight}), and (\ref{eq:gen_total_cost3}). This revised formulation models the inherent flexibility of GED and enhances node alignment efficiency:
\begin{align}
&c_{q,d} = \text{Estimator}(\text{cost}_{{q,d}}, \Psi(\Phi(q), \Phi(d))),\label{eq:gen_gnn_cost_cor}\\
& w_{q,d} = \text{WeightCompute}\big(c_{q,d},\mathbf{A}_Q,\mathbf{A}_D\big),\label{eq:gen_weight_cor}\\
&\textbf{c}(Q,D) = \sum_{q\in \mathcal{Q},d\in \mathcal{D}} w_{q,d} \cdot c_{q,d}. \label{eq:gen_total_cost3_cor}
\end{align}
Equation~(\ref{eq:gen_gnn_cost_cor}) estimates the expense of matching a candidate node pair $(q,d)$ by jointly considering the node-level difference $dist_{q,d} = \Psi(\Phi(q), \Phi(d))$, which encodes intrinsic difference and local structural context, and the corresponding operation cost vector $\text{cost}_{q,d}$. This design minimizes the \textit{Expense Gap} $\|\Delta C\|_F$ by coupling operation costs and intrinsic differences with node matching expense. To minimize the \textit{Alignment Gap} $\|\Delta W\|_F$, the second equation computes weight $w_{q,d}$ for each candidate match by incorporating both intra-graph and inter-graph dependencies. 
Finally, Equation~(\ref{eq:gen_total_cost3_cor}) aggregates all weighted node matching expenses to approximate the total edit expense $\textbf{c}(Q,D)$ between graphs. Building on this formulation, we introduce the Graph Edit Network (\name) framework which concurrently minimize the Cost and Alignment Gaps in Eq. (\ref{eq:bound}). We deliberately adopt a minimalistic implementation of $\text{Estimator}(\cdot,\cdot)$ and $\text{WeightCompute}(\cdot)$ to demonstrate that even a simple, GED-principled revision can overcome the limitations of existing approximation paradigms.

\subsection{\name\ Overview}

The overview of \name\ is illustrated in Figure \ref{fig:overview}. Given a query-data graph pair and the associated atomic edit operation cost vector, \name\ first embeds both graphs into a shared representation space using a \emph{shared GNN encoder} $\Phi(\cdot)$, ensuring semantic alignment between node embeddings. Guided by the modified Equations~(\ref{eq:gen_gnn_cost_cor}), (\ref{eq:gen_weight_cor}) and (\ref{eq:gen_total_cost3_cor}), \name\ introduces two key modules: \emph{cost-aware matching expense estimation} $\text{Estimator}(\cdot, \cdot)$ and \emph{one-step node alignment with global guidance}  $\text{WeightCompute}(\cdot)$. These components work in tandem to estimate the node matching weights $w$. Specifically, $\text{Estimator}(\cdot, \cdot)$ integrates operation costs with embedding distances to estimate node matching expenses under the current cost setting. $\text{WeightCompute}(\cdot)$ evaluates each candidate match's relative impact on the overall edit expense and propagates impacts within and between graphs to capture both intra- and inter-graph match dependencies. The resulting global guidance signal is directly used to compute matching weights, and the weighted sum of node matching costs provides an approximation of GED.

\subsection{Cost-Aware Matching Expense Estimation}

Each node $v$ is initially represented by a one-hot encoding of its label $\mathbf{x}_v^0 \in \mathbb{R}^{|\mathcal{X}|}$, where $|\mathcal{X}|$ denotes the number of label types. In unlabeled graphs, all nodes can be treated as sharing the same label. For each graph, \name\ employs a $k$-layer GNN encoder to generate node embeddings $\mathbf{X}^k$ capturing $k$-hop local structure. The layer-wise update for node $v$ is:
\begin{align}
    \mathbf{x}^{l}_v = \text{Update}\big(\mathbf{x}^{l-1}_v, \text{Aggr}(\mathbf{x}^{l-1}_u : \mathbf{A}_{v,u} = 1)\big),
\end{align}
where $\text{Aggr}(\cdot)$ is a aggregation function that gather information from direct neighbors of $v$, and $\text{Update}(\cdot,\cdot)$ combines the $v$'s previous representation with aggregated neighbor information. They are both permutation-invariant operators (e.g., Sum, Mean, Max). The final $k$-hop embedding $\mathbf{x}^k_v$ is projected via a MLP to dimension $H$.

To ensure equal node counts, \name\ pads the smaller graph with virtual node features initialized to zeros. Given initial node features $\mathbf{X}^0$, high-order embeddings $\mathbf{X}^k$. For each node $v$, \name\ concatenate its initial feature and embedding to encodes both label and structural information as follows: 
\begin{equation}
\mathbf{h}_v = \text{Concat}(\mathbf{x}^0_v, \mathbf{x}^k_v),
\end{equation}
where $\mathbf{h}_v\in\mathbb{R}^{|\mathcal{X}| + H}$. Therefore, \name\ computes the distance between $q \in \mathcal{V}_Q$ and $d \in \mathcal{V}_D$ based on their embeddings, enabling $k$-hop contextual estimation without explicitly comparing neighborhood structures.
\begin{equation}
dist_{q,d} = \mathbf{h}_q - \mathbf{h}_d,
\end{equation}
where the computation of $dist{q,d}$ adopts an asymmetric formulation to accommodate both uniform and non-uniform atomic operation costs.

To account for operation costs, $\text{cost}_{q,d}$ is provided as input for cost-aware matching expense estimation:
\begin{equation}
c_{q,d} = \text{MLP}\big(\text{Concat}(\text{cost}_{q,d}, dist_{q,d})\big), \label{eq:estimator}
\end{equation}
The use of MLP aligns with estimator design in prior fine-grained works, which serves as a functional approximator \cite{HORNIK1989359} for the cost-aware expense matrix. Specifically, its fully connected and non-linear layers enable the mapping of relevant feature dimensions to specific edit operations and approximate the ``switching'' of operations, minimizing the Expense Gap term in Eq. (\ref{eq:bound}) prior to alignment.

\subsection{One-Step Node Alignment with Global
Guidance}
The alignment stage is designed as a relaxed neural mechanism inspired by the structural gradient formulation of Quadratic Assignment Problem (QAP) \cite{qap}, aiming to capture the notion of the Alignment Gap. It captures both node-level similarities (first-order term) and structural dependencies (second-order term).

\subsubsection{Global Guidance Initialization}

Node matching expenses can vary widely, and selecting matches based solely on individual expenses may yield locally optimal but globally suboptimal alignments. To address this, \name\ initializes a global guidance signal by normalizing node-level matching expense vectors into global cost ratios:
\begin{equation}
\gamma_{q,d} = \frac{\exp(c_{q,d})}{\sum_{q \in \mathcal{V}_Q, d \in \mathcal{V}_D} \exp(c_{q,d})},
\end{equation}
where the normalization is applied element-wise across the embedding dimensions of $c_{q,d} \in \mathbb{R}^H$. The resulting $\gamma_{q,d}$ captures the relative global impact of each candidate match along each embedding channel. Node-level impact vectors are then obtained by summation over candidate matches:
\begin{align}
\gamma_q = \sum_{i \in \mathcal{V}_D} \gamma_{q,i}, \quad
\gamma_d = \sum_{i \in \mathcal{V}_Q} \gamma_{i,d},
\end{align}
where higher values indicate that accurately matching the node is critical for minimizing the overall edit expense, which can be interpreted as analogous to the first-order component in the QAP gradient.

\subsubsection{Impact Propagation}

To capture intra-graph dependencies, \name\ propagates node impact across neighborhoods, capturing match interactions in a manner reminiscent of the second-order term in the QAP objective.
\begin{equation}
\gamma_{\mathcal{N}(v)} = \text{MLP}\Big(\text{Sum}(\gamma_v, \sum_{\mathbf{A}_{v,u}=1} \gamma_u)\Big),
\end{equation}
and incorporates residual node embeddings to retain node-specific semantics and stabilize propagation, enhancing the discriminative power of the subsequent weight computation:
\begin{equation}
\gamma_{\mathcal{N}(v)} \leftarrow \text{MLP}\Big(\text{Concat}(\gamma_{\mathcal{N}(v)}, \mathbf{h}_v)\Big).
\end{equation}
Cross-graph matching then balances the inter-graph contributions of matches by concatenating the propagated neighborhood impact from both graphs. The final weight for each candidate pair $(q,d)$ is computed as
\begin{equation}
w_{q,d} = \text{MLP}\Big(\text{Concat}(\gamma_{\mathcal{N}(q)}, \gamma_{\mathcal{N}(d)})\Big),
\end{equation}
producing a globally informed, one-step alignment that accounts for intrinsic node differences, operation costs, and match dependencies. All operations leverage sparse scatter computations for GPU efficiency. This global guidance serves as a lightweight structural prior, capturing essential dependencies among node matches without explicitly modeling all GED dependencies.

\subsection{Prediction and Training Objective}

\subsubsection{Prediction}

Given candidate weights $w_{q,d}$, node-level edit expenses are computed as
\begin{align}
c_q &= \sum_{i \in \mathcal{V}_D} w_{q,i} \cdot c_{q,i},\\
c_d &= \sum_{i \in \mathcal{V}_Q} w_{i,d} \cdot c_{i,d},
\end{align}
and aggregated to obtain graph-level expenses:
\begin{align}
\textbf{c}(Q,D)_{f_Q} &= \sum_{q \in \mathcal{V}_Q} c_q,\\
\textbf{c}(Q,D)_{f_D} &= \sum_{d \in \mathcal{V}_D} c_d.
\end{align}
The final GED estimate is obtained via
\begin{equation}
\text{GED}(Q,D) = \text{Linear}\Big(\text{Mean}(\textbf{c}(Q,D)_{f_Q}, \textbf{c}(Q,D)_{f_D})\Big),
\end{equation}
implicitly encouraging cross-graph consistency in candidate selection.
\subsubsection{Training Objective}

Let $\hat{y}$ denote the predicted GED and $y$ the ground-truth GED. We adopt the Huber loss~\cite{huber}:
\begin{equation}
\mathcal{L}(\hat{y}, y) =
\begin{cases}
\frac{1}{2}(y - \hat{y})^2 & \text{if } |y-\hat{y}| \leq \delta \\
\delta \left( |y-\hat{y}| - \frac{1}{2}\delta \right) & \text{otherwise,}
\end{cases}
\end{equation}
where $\delta$ controls the transition between quadratic and linear regimes, providing robustness against outliers compared to standard MSE.

\subsection{Complexity Analysis}
The complexity of the \emph{GNN Encoder} is $\mathcal{O}(k |\mathcal{E}| d_{\text{hidden}})$, where $k$ is the number of GNN layers and $\mathcal{E}$ is the set of edges. The complexity of \emph{Cost-Aware Matching Expense Computation} is $\mathcal{O}(\max(|\mathcal{V}_S|, |\mathcal{V}_T|)^2\cdot d)$. The \emph{One-Step Node Alignment with Global Guidance} introduces an additional complexity of $\mathcal{O}(2(\max(|\mathcal{V}_S|, |\mathcal{V}_T|)^2 d + |\mathcal{E}|d))$.

\begin{table*}[ht]
\caption{Statistics of datasets.}
\resizebox{\textwidth}{!}{ \renewcommand{\arraystretch}{1.0}
\centering
\begin{tabular} {c|cccccccccc}
\toprule
\textbf{Dataset} &
\textbf{\#Graphs} &
\textbf{\#Pairs} &
\textbf{\#Testing Pairs} &
\textbf{\#Labels} &
\textbf{Avg. \#nodes} &
\textbf{Avg. \#edges} &
\textbf{Max. \#nodes} &
\textbf{Max. \#edges} &
\textbf{Min. \#nodes} &
\textbf{Min. \#edges} \\ 
\midrule
\textbf{AIDS700nef} &
700 &
490K &
78,4K &
29 &
8.90 &
8.80 &
10.00 &
14.00 &
2.00 &
1.00 \\ 

\textbf{LINUX} &
1000 &
1M &
160k &
1 &
7.58 &
6.94 &
10.00 &
13.00 &
4.00 &
3.00 \\ 
\textbf{IMDBMulti} &
1500 &
2.25M &
360k &
1 &
13.00 &
65.94 &
89.00 &
1467.00 &
7.00 &
12.00 \\ 
\textbf{Synthetic} &
51,200 &
25,6 K &
- &
1 &
230.24 &
1789.02 &
400.00 &
8552.00 &
100.00 &
100.00 \\ 

\textbf{[100, 200]} &
12,800 &
6.4K &
- &
1 &
148.61 &
1328.51 &
200.00 &
5806.00 &
100.00 &
196.00 \\ 

\textbf{[200, 300]} &
12,800 &
6.4K &
- &
1 &
247.45 &
1979.85 &
300.00 &
7638.00 &
200.00 &
396.00 \\ 

\textbf{[300, 400]} &
12,800 &
6.4K &
- &
1 &
348.14 &
2295.15 &
400.00 &
8083.00 &
300.00 &
596.00 \\ 
\bottomrule
\end{tabular}}
\centering \label{tab:statistic}
\end{table*}

\section{Evaluations}
This section evaluates \name\ against twelve baselines, including two conventional inexact GED algorithms, on three widely used GED datasets and synthetic graphs. The evaluation focuses on the following aspects:
\begin{enumerate}
\item \emph{Effectiveness}: How accurately \name\ computes GED under a uniform cost setting, where all atomic operations have equal costs;
\item \emph{Time Efficiency}: How efficiently \name\ performs in terms of training time, inference throughput, and peak memory consumption;
\item \emph{Case Study}: How well \name\ ranks graphs in representative query cases;
\item \emph{Handling Non-uniform and Mixed Costs}: Whether \name\ and the baselines can work under non-uniform and mixed cost settings;
\item \emph{Ablation Study}: How effective the proposed global guidance is;
\item \emph{Hyperparameter Sensitivity}: How sensitive \name\ is to variations in the GNN backbones, number of layers, and embedding dimensionality;
\item \emph{Performance on Larger Synthetic Graphs}: How \name\ and the baselines perform on larger synthetic graphs.
\end{enumerate}

\subsection{Experimental Setup} 
\subsubsection{Dataset}
Following previous learning-based works published in database venues \cite{gedgnn,noah}, we conduct experiments on three widely used GED datasets to ensure fair comparison:
\begin{itemize}
    \item \textbf{AIDS700nef}: 700 node-labeled chemical compound structures, each containing at most ten nodes;
    \item \textbf{LINUX}: 1,000 unlabeled Program Dependence Graphs (PDGs) generated from the Linux kernel, each with at most ten nodes;
    \item \textbf{IMDBMulti}: 1,500 unlabeled ego-networks of movie actors and actresses, each containing up to 89 nodes.
\end{itemize}
For these datasets, the graphs and their exact or approximate GED values (for IMDBMulti) under a uniform cost setting are collected and provided by \cite{simgnn}, yielding a large number of graph pairs. Using the same graphs in AIDS700nef, we additionally compute GED values under two non-uniform cost settings with the A*-based algorithm \cite{A*GED}. Note that exact GED computation already becomes challenging for graphs with 16 nodes \cite{exact}.

\begin{figure}[!htb]
    \centering
    \begin{subfigure}{0.49\linewidth}
    \centering
    \includegraphics[width=\linewidth]{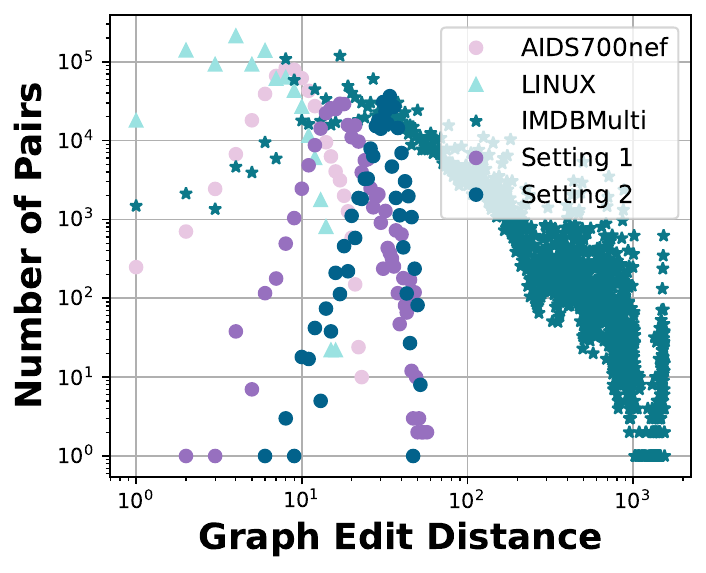}
    \end{subfigure}
    \begin{subfigure}{0.49\linewidth}
    \centering
    \includegraphics[width=\linewidth]{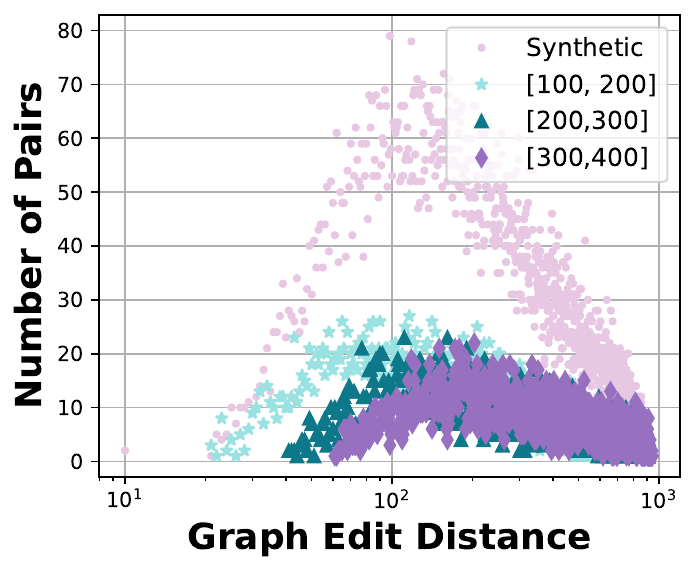}
    \end{subfigure}
    \caption{The GED distributions of datasets.}
    \label{fig:dist_main}
\end{figure}

Additionally, we generate synthetic graphs following \cite{gedgnn}, including Erdős–Rényi, power-law, and Watts–Strogatz small-world graphs, with node sizes ranging from 100 to 400. These graphs serve as extreme cases to further evaluate the effectiveness of the models under the impact of graph size. For each synthetic graph, we apply $\Delta$ distinct operations to generate a corresponding query graph, where $\Delta$ ranges from 10 to 30 percent of the number of edges. Following previous work \cite{gmn,tagsim,gedgnn}, we regard $\Delta$ as the ground-truth GED for synthetic graph pairs. In total, we generate 51,200 graphs for training and another 38,400 graphs for testing, divided into three node size groups: $[100,200]$, $[200,300]$, and $[300,400]$. The detailed graph statistics and the GED distributions are provided in Table~\ref{tab:statistic} and Figure~\ref{fig:dist_main}, respectively. Notably, although the graphs are of moderate size, the GED values can reach the order of $10^3$, highlighting the substantial challenge of accurately predicting GED in these settings.

\subsubsection{Baselines}

We include representative, relevant and state-of-the-art learning-based GED models as baselines: (1) \emph{Graph-level methods}: GMN-emb \cite{gmn}, TaGSim \cite{tagsim}, Greed \cite{greed}, and Eric \cite{eric}; (2) \emph{Node-level methods}: GMN-match \cite{gmn}, SimGNN \cite{simgnn}, GraphSim \cite{graphsim}, GOTSim \cite{gotsim}, GEDGNN \cite{gedgnn}, GraphEdX \cite{graphedx} and GEDIOT \cite{sigmod1}. In addition, we compare with two classical and recent inexact GED algorithms, namely the Volgenant–Jonker (VJ) algorithm \cite{vj}, Beam search \cite{beam} and FGWAlign \cite{vldb1}. The Hungarian algorithm \cite{hun} and MATA* \cite{mata*} are not included, as they follow a paradigm already represented by the included baselines.

\begin{itemize}
\item \textbf{GMN-emb} \cite{gmn}: A representative graph-level method that generates graph representations by applying gated aggregation over node representations.
\item \textbf{TaGSim} \cite{tagsim}: A graph-level method that models operation types and predicts the count of each operation type.
\item \textbf{Greed} \cite{greed}: A representative method that incorporates inductive bias through the triangle inequality of GED to enhance estimation.
\item \textbf{Eric} \cite{eric}: A state-of-the-art graph-level method that enforces a node–graph alignment constraint during training in an unsupervised manner, and decouples this module during inference to improve efficiency.
\item \textbf{GMN-match} \cite{gmn}: A representative and relevant node-level method that updates node representations by propagating within and across graphs.
\item \textbf{SimGNN} \cite{simgnn}: A node-level method that computes histogram features from pairwise node similarity scores.
\item \textbf{GraphSim} \cite{graphsim}: A representative node-level method that reorders the similarity matrix using BFS traversal to capture local matching affinities.
\item \textbf{GOTSim} \cite{gotsim}: A representative node-level method that explicitly establishes node alignments using a CPU-based linear assignment solver.
\item \textbf{GEDGNN} \cite{gedgnn}: A representative node-level method that explicitly builds node alignments by optimizing graph transport costs, and uses the trained model to guide the search process of exact GED algorithms.
\item \textbf{GraphEdX} \cite{graphedx}: A relevant and state-of-the-art node-level method, which defines a family of distance measures for different operation types and reweighs alignments using operation costs.
\item \textbf{GEDIOT} \cite{sigmod1}: Most recent node-level method, which is similar to GEDGNN, where the latter derives the mapping matrix from node embedding distances, while the former extracts it from node alignments processed through Sinkhorn.
\item \textbf{VJ} \cite{vj}: The Volgenant–Jonker algorithm solves linear sum assignment via greedy bipartite matching to minimize node-level costs.
\item \textbf{Beam} \cite{beam}: Beam search adopts a best-first strategy, expanding partial solutions with the lowest-cost matches at each step to approximate GED.
\item \textbf{FGWAlign} \cite{vldb1}: FGWAlign leverages Sinkhorn algorithm to align nodes and compute node edit expenses, then generate edit path to consider edge edit expenses.
\end{itemize}

We use the official implementations and hyperparameter settings provided by the respective authors whenever available. The implementations of VJ and Beam are sourced from \cite{noah}. For models that predict normalized GED values, we remove their final sigmoid layer to ensure fair comparison.

\subsubsection{Implementation Details}
We adopt a 5-layer Graph Isomorphism Network (GIN) \cite{GIN} with jump connections as the GNN encoder. The hidden and output dimensions of the GIN are set to 64, and a final MLP reduces the node embeddings to 32 dimensions. All MLPs in this work employ layer normalization and ReLU activations, with an output dimensionality of 32. Optimization is performed using the Adam optimizer with a fixed learning rate of 0.001.

\subsubsection{Experimental Protocol} 
All models are trained on an NVIDIA A6000 GPU with an AMD EPYC 7282 CPU. The batch size is set to 128 graph pairs for training and 2048 for validation and testing. Due to the large number of possible graph pairs, each epoch consists of 100 iterations, except for synthetic datasets where the number of pairs is limited. Each model is trained for 10 warm-up epochs and evaluated on the validation set at the end of each epoch. Early stopping with a patience of 50 is applied to prevent overfitting while ensuring sufficient training, terminating training if the validation loss does not improve for 50 consecutive validation steps.

\subsubsection{Evaluation Metrics}
We evaluate models in terms of prediction error and ranking ability. Prediction error measures the point-wise difference between predicted and ground-truth GED values. Ranking ability is assessed by simulating retrieval scenarios in graph databases, where training and validation graphs serve as data graphs and test graphs act as queries. During testing, each model computes the GED between a query and all data graphs, returning the most relevant graphs with the lowest predicted GED. For prediction error, we use \emph{Root Mean Square Error (RMSE)} and \emph{Mean Absolute Error (MAE)}. RMSE, like MSE, is sensitive to outliers but reports errors in the original units, improving interpretability, whereas MAE captures general deviations. For ranking performance, we employ \emph{Spearman's Rank Correlation Coefficient ($\rho$)}, \emph{Kendall's Rank Correlation Coefficient ($\tau$)}, and \emph{Precision at $k$ ($p@k$)}. $\rho$ is sensitive to the magnitude of rank differences, while $\tau$ evaluates the relative ordering of pairs. $p@k$ measures the overlap between the predicted top-$k$ data graphs and the ground-truth top-$k$ results. Rankings are computed by sorting all data graphs for each query based on both predicted and ground-truth GEDs in ascending order. Results are reported as the mean and standard deviation over five runs.

\begin{table}
\centering
\caption{Results on AIDS700nef. We mark the \textbf{best} and the \underline{second} models.}\label{tab:aids_res}
\resizebox{\linewidth}{!}{ \renewcommand{\arraystretch}{1.0}
\centering
\begin{tabular}{l|ccccc}
\toprule
\textbf{Method} &
\textbf{RMSE} $\downarrow$ &
\textbf{MAE} $\downarrow$ &
\textbf{$\rho$} $\uparrow$ &
\textbf{$\tau$} $\uparrow$ &
\textbf{p@10} $\uparrow$ \\ 
\midrule
\textbf{VJ} &
$9.278_{\pm\text{0.000}}$ &
$8.709_{\pm\text{0.000}}$ &
$0.422_{\pm\text{0.000}}$ &
$0.296_{\pm\text{0.000}}$ &
$0.269_{\pm\text{0.000}}$ \\
\textbf{Beam} &
$5.111_{\pm\text{0.000}}$ &
$4.380_{\pm\text{0.000}}$ &
$0.486_{\pm\text{0.000}}$ &
$0.345_{\pm\text{0.000}}$ &
$0.371_{\pm\text{0.000}}$ \\
\textbf{FGWAlign} &
$2.006_{\pm\text{0.000}}$ &
$1.426_{\pm\text{0.000}}$ &
$\underline{0.699_{\pm\text{0.000}}}$ &
$\underline{0.525_{\pm\text{0.000}}}$ &
$\underline{0.469_{\pm\text{0.000}}}$ \\
\midrule
\textbf{SimGNN} &
$1.036_{\pm\text{0.040}}$ &
$1.075_{\pm\text{0.084}}$ &
$0.847_{\pm\text{0.008}}$ &
$0.670_{\pm\text{0.008}}$ &
$0.897_{\pm\text{0.005}}$ \\
\textbf{GMN-match} &
$1.068_{\pm\text{0.032}}$ &
$0.811_{\pm\text{0.021}}$ &
$0.873_{\pm\text{0.006}}$ &
$0.701_{\pm\text{0.008}}$ &
$0.899_{\pm\text{0.004}}$ \\
\textbf{GMN-emb}  &
$1.645_{\pm\text{0.081}}$ &
$1.145_{\pm\text{0.033}}$ &
$0.854_{\pm\text{0.006}}$ &
$0.679_{\pm\text{0.006}}$ &
$0.793_{\pm\text{0.025}}$ \\
\textbf{GraphSim} &
$2.017_{\pm\text{0.793}}$ &
$1.519_{\pm\text{0.580}}$ &
$0.382_{\pm\text{0.381}}$ &
$0.298_{\pm\text{0.306}}$ &
$0.365_{\pm\text{0.447}}$ \\
\textbf{GOTSim} &
$1.584_{\pm\text{0.347}}$ &
$1.198_{\pm\text{0.249}}$ &
$0.691_{\pm\text{0.183}}$ &
$0.525_{\pm\text{0.149}}$ &
$0.783_{\pm\text{0.058}}$ \\
\textbf{TaGSim} &
$1.113_{\pm\text{0.058}}$ &
$0.844_{\pm\text{0.039}}$ &
$0.841_{\pm\text{0.012}}$ &
$0.663_{\pm\text{0.014}}$ &
$0.896_{\pm\text{0.009}}$ \\
\textbf{Eric} &
$1.259_{\pm\text{0.316}}$ &
$0.971_{\pm\text{0.228}}$ &
$0.782_{\pm\text{0.097}}$ &
$0.605_{\pm\text{0.095}}$ &
$0.836_{\pm\text{0.111}}$ \\
\textbf{Greed} &
$0.900_{\pm\text{0.016}}$ &
$0.705_{\pm\text{0.012}}$ &
$0.885_{\pm\text{0.001}}$ &
$0.715_{\pm\text{0.002}}$ &
$0.930_{\pm\text{0.005}}$ \\
\textbf{GEDGNN} &
$1.607_{\pm\text{0.618}}$ &
$1.148_{\pm\text{0.354}}$ &
$0.747_{\pm\text{0.114}}$ &
$0.577_{\pm\text{0.104}}$ &
$0.739_{\pm\text{0.155}}$ \\
\textbf{GraphEdX} &
$\textbf{0.735}_{\pm\textbf{0.015}}$ &
$\underline{0.581_{\pm\text{0.011}}}$ &
$0.910_{\pm\text{0.003}}$ &
$0.749_{\pm\text{0.004}}$ &
$0.951_{\pm\text{0.003}}$ \\
\textbf{GEDIOT} &
$0.776_{\pm\text{0.019}}$ &
$0.603_{\pm\text{0.013}}$ &
$\underline{0.911_{\pm\text{0.002}}}$ &
$\underline{0.750_{\pm\text{0.003}}}$ &
$\underline{0.954_{\pm\text{0.005}}}$ \\
\midrule
\textbf{Ours (\name)} &
$\underline{0.738_{\pm\text{0.007}}}$ &
$\textbf{0.577}_{\pm\textbf{0.005}}$ &
$\textbf{0.915}_{\pm\textbf{0.002}}$ &
$\textbf{0.756}_{\pm\textbf{0.003}}$ &
$\textbf{0.963}_{\pm\textbf{0.003}}$ \\
\bottomrule
\end{tabular}}
\end{table}

\begin{table}
\centering
\caption{Results on LINUX.}\label{tab:linux_res}
\resizebox{\linewidth}{!}{ \renewcommand{\arraystretch}{1.0}
\centering
\begin{tabular}{l|ccccc}
\toprule
\textbf{Method} &
\textbf{RMSE} $\downarrow$ &
\textbf{MAE} $\downarrow$ &
\textbf{$\rho$} $\uparrow$ &
\textbf{$\tau$} $\uparrow$ &
\textbf{p@10} $\uparrow$ \\ 
\midrule
\textbf{VJ} &
$3.407_{\pm\text{0.000}}$ &
$2.692_{\pm\text{0.000}}$ &
$0.728_{\pm\text{0.000}}$ &
$0.574_{\pm\text{0.000}}$ &
$0.934_{\pm\text{0.000}}$ \\
\textbf{Beam} &
$3.315_{\pm\text{0.000}}$ &
$2.841_{\pm\text{0.000}}$ &
$0.734_{\pm\text{0.000}}$ &
$0.575_{\pm\text{0.000}}$ &
$0.445_{\pm\text{0.000}}$ \\
\textbf{FGWAlign} &
$2.002_{\pm\text{0.000}}$ &
$1.588_{\pm\text{0.000}}$ &
$\underline{0.727_{\pm\text{0.000}}}$ &
$\underline{0.569_{\pm\text{0.000}}}$ &
$\underline{0.604_{\pm\text{0.000}}}$ \\
\midrule
\textbf{SimGNN} & 
$0.621_{\pm\text{0.076}}$ &
$0.434_{\pm\text{0.066}}$ &
$0.919_{\pm\text{0.009}}$ &
$0.760_{\pm\text{0.012}}$ &
$0.869_{\pm\text{0.065 }}$ \\
\textbf{GMN-match} &
$\underline{0.267_{\pm\text{0.007}}}$ &
$0.126_{\pm\text{0.008}}$ &
$\underline{0.946_{\pm\text{0.001}}}$ &
$0.805_{\pm\text{0.002}}$ &
$0.967_{\pm\text{0.009}}$ \\
\textbf{GMN-emb}  &
$0.334_{\pm\text{0.033}}$ &
$0.174_{\pm\text{0.010}}$ &
$0.944_{\pm\text{0.001}}$ &
$0.803_{\pm\text{0.002}}$ &
$0.974_{\pm\text{0.014}}$ \\
\textbf{GraphSim} &
$2.283_{\pm\text{1.008}}$ &
$1.721_{\pm\text{0.800}}$ &
$0.206_{\pm\text{0.371}}$ &
$0.175_{\pm\text{0.319}}$ &
$0.276_{\pm\text{0.359}}$ \\
\textbf{GOTSim} &
$1.560_{\pm\text{0.621}}$ &
$1.242_{\pm\text{0.460}}$ &
$0.666_{\pm\text{0.363}}$ &
$0.528_{\pm\text{0.281}}$ &
$0.550_{\pm\text{0.066}}$ \\
\textbf{TaGSim} &
$0.678_{\pm\text{0.066}}$ &
$0.438_{\pm\text{0.051}}$ &
$0.913_{\pm\text{0.010}}$ &
$0.758_{\pm\text{0.012}}$ &
$0.909_{\pm\text{0.029}}$ \\
\textbf{Eric} &
$0.271_{\pm\text{0.130}}$ &
$\underline{0.120_{\pm\text{0.104}}}$ &
$0.944_{\pm\text{0.007}}$ &
$0.803_{\pm\text{0.012}}$ &
$0.973_{\pm\text{0.027}}$ \\
\textbf{Greed} &
$0.441_{\pm\text{0.003}}$ &
$0.327_{\pm\text{0.002}}$ &
$0.939_{\pm\text{0.001}}$ &
$0.791_{\pm\text{0.001}}$ &
$0.960_{\pm\text{0.002}}$ \\
\textbf{GEDGNN} &
$0.313_{\pm\text{0.098}}$ &
$0.188_{\pm\text{0.071}}$ &
$0.945_{\pm\text{0.005}}$ &
$\underline{0.806_{\pm\text{0.007}}}$ &
$\underline{0.993_{\pm\text{0.005}}}$ \\
\textbf{GraphEdX} &
$0.378_{\pm\text{0.135}}$ &
$0.222_{\pm\text{0.081}}$ &
$0.941_{\pm\text{0.006}}$ &
$0.797_{\pm\text{0.011}}$ &
$0.973_{\pm\text{0.021}}$ \\
\textbf{GEDIOT} &
$0.390_{\pm\text{0.033}}$ &
$0.168_{\pm\text{0.013}}$ &
$0.942_{\pm\text{0.002}}$ &
$0.805_{\pm\text{0.001}}$ &
$0.991_{\pm\text{0.005}}$ \\
\midrule
\textbf{Ours (\name)} &
$\textbf{0.166}_{\pm\textbf{0.009}}$ &
$\textbf{0.084}_{\pm\textbf{0.008}}$ &
$\textbf{0.948}_{\pm\textbf{0.000}}$ &
$\textbf{0.810}_{\pm\textbf{0.001}}$ &
$\textbf{0.995}_{\pm\textbf{0.004}}$ \\
\bottomrule
\end{tabular}}
\vspace{-0.5cm}
\end{table}
\begin{table}
\centering
\caption{Results on IMDBMulti. 'OOM' indicates out of memory errors.}\label{tab:imdb_res}
\resizebox{\linewidth}{!}{ \renewcommand{\arraystretch}{1.0}
\centering
\begin{tabular}{l|ccccc}
\toprule
\textbf{Method} &
\textbf{RMSE} $\downarrow$ &
\textbf{MAE} $\downarrow$ &
\textbf{$\rho$} $\uparrow$ &
\textbf{$\tau$} $\uparrow$ &
\textbf{p@10} $\uparrow$ \\ 
\midrule
\textbf{SimGNN} & 
$13.721_{\pm\text{2.163}}$ &
$6.863_{\pm\text{1.161}}$ &
$0.899_{\pm\text{0.006}}$ &
$0.781_{\pm\text{0.006}}$ &
$0.978_{\pm\text{0.001}}$ \\
\textbf{GMN-match} &
$\underline{8.401_{\pm\text{2.061}}}$ &
$\underline{3.694_{\pm\text{0.925}}}$ &
$\textbf{0.938}_{\pm\textbf{0.007}}$ &
$\textbf{0.843}_{\pm\textbf{0.017}}$ &
$\underline{0.985_{\pm\text{0.004}}}$ \\
\textbf{GMN-emb}  &
$12.340_{\pm\text{1.403}}$ &
$5.296_{\pm\text{0.707}}$ &
$0.926_{\pm\text{0.005}}$ &
$0.820_{\pm\text{0.010}}$ &
$0.966_{\pm\text{0.034}}$ \\
\textbf{GraphSim} &
$159.571_{\pm\text{0.012}}$ &
$70.917_{\pm\text{0.002}}$ &
$0.003_{\pm\text{0.000}}$ &
$0.002_{\pm\text{0.000}}$ &
$0.001_{\pm\text{0.000}}$ \\
\textbf{GOTSim} &
$129.623_{\pm\text{13.795}}$ &
$52.248_{\pm\text{7.199}}$ &
$0.611_{\pm\text{0.250}}$ &
$0.492_{\pm\text{0.208}}$ &
$0.474_{\pm\text{0.201}}$ \\
\textbf{TaGSim} &
$11.565_{\pm\text{3.310}}$ &
$5.451_{\pm\text{1.340}}$ &
$0.873_{\pm\text{0.053}}$ &
$0.764_{\pm\text{0.059}}$ &
$0.979_{\pm\text{0.008}}$ \\
\textbf{Eric} &
$9.866_{\pm\text{0.662}}$ &
$4.462_{\pm\text{0.292}}$ &
$0.932_{\pm\text{0.005}}$ &
$0.834_{\pm\text{0.008}}$ &
$0.977_{\pm\text{0.006}}$ \\
\textbf{Greed} &
$19.974_{\pm\text{20.988}}$ &
$6.948_{\pm\text{5.880}}$ &
$0.890_{\pm\text{0.039}}$ &
$0.778_{\pm\text{0.047}}$ &
$0.837_{\pm\text{0.303}}$ \\
\textbf{GEDGNN} &
$15.743_{\pm\text{9.533}}$ &
$7.070_{\pm\text{3.471}}$ &
$0.897_{\pm\text{0.036}}$ &
$0.785_{\pm\text{0.046}}$ &
$0.934_{\pm\text{0.076}}$ \\
\textbf{GraphEdX} &
OOM &
OOM &
OOM &
OOM &
OOM \\
\textbf{GEDIOT} &
$40.834_{\pm\text{34.978}}$ &
$16.476_{\pm\text{13.861}}$ &
$0.794_{\pm\text{0.172}}$ &
$0.684_{\pm\text{0.173}}$ &
$0.887_{\pm\text{0.091}}$ \\
\midrule
\textbf{Ours (\name)} &
$\textbf{7.531}_{\pm\textbf{0.293}}$ &
$\textbf{3.492}_{\pm\textbf{0.128}}$ &
$\underline{0.937_{\pm\text{0.002}}}$ &
$\underline{0.839_{\pm\text{0.002}}}$ &
$\textbf{0.987}_{\pm\textbf{0.001}}$ \\
\bottomrule
\end{tabular}}
\end{table}
\subsection{Results} \subsubsection{Effectiveness}
We compare the performance of \name\ with all baselines under a uniform cost setting, where all atomic operation costs are set to 1. Results for VJ and Beam on IMDBMulti are excluded because the training targets of this dataset correspond to the smallest distances produced by these algorithms. As shown in Table~\ref{tab:aids_res}, \ref{tab:linux_res}, and \ref{tab:imdb_res}, \name\ consistently ranks first or second across all metrics and datasets, demonstrating strong overall performance. In addition, \name\ exhibits the lowest variance across runs among learning-based baselines, indicating more stable training and prediction. \name\ achieves up to a 37.8\% reduction in RMSE and a 30\% reduction in MAE on LINUX, and improves ranking performance by up to 4.1\%, with only minor reductions of 0.4\% in RMSE on AIDS700nef and 0.4\% in Kendall’s $\tau$ on IMDBMulti.

Learning-based methods generally outperform conventional ones, as they can encode richer structural information through deep neural architectures. Among learning-based approaches, graph-level methods (e.g., Greed, Eric) outperform most node-level methods, indicating no inherent superiority of one paradigm over the other under uniform costs. \name\ clearly surpasses most node-level approaches, which often use isolated node distances to guide alignment refinement. In contrast, \name\ employs a global guidance mechanism that accounts for relative matching expenses as well as inter- and intra-graph dependencies of candidate matches, enabling more informed one-step alignments.

\begin{figure}[!htb]
    
    \centering
    \includegraphics[width=\linewidth]{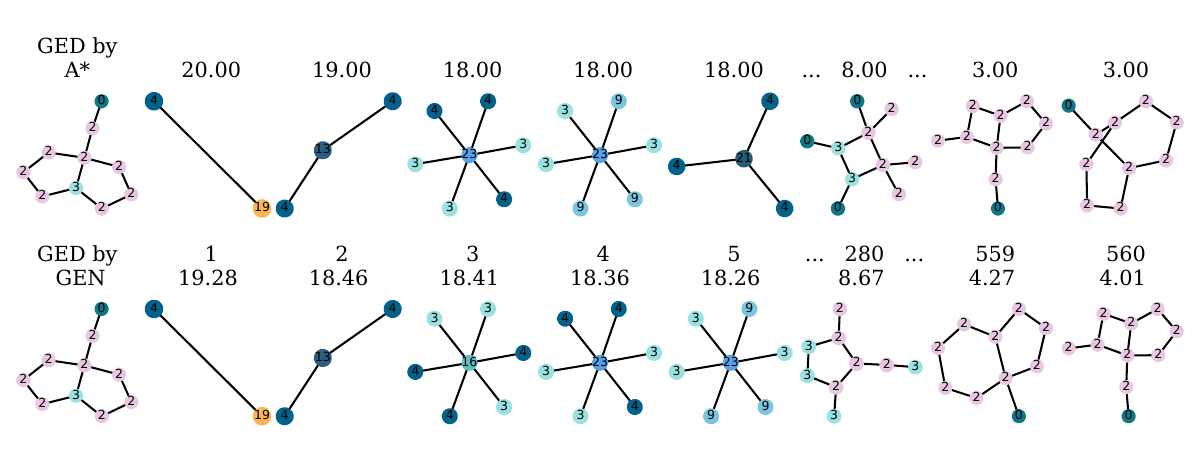}\\
    
    \includegraphics[width=\linewidth]{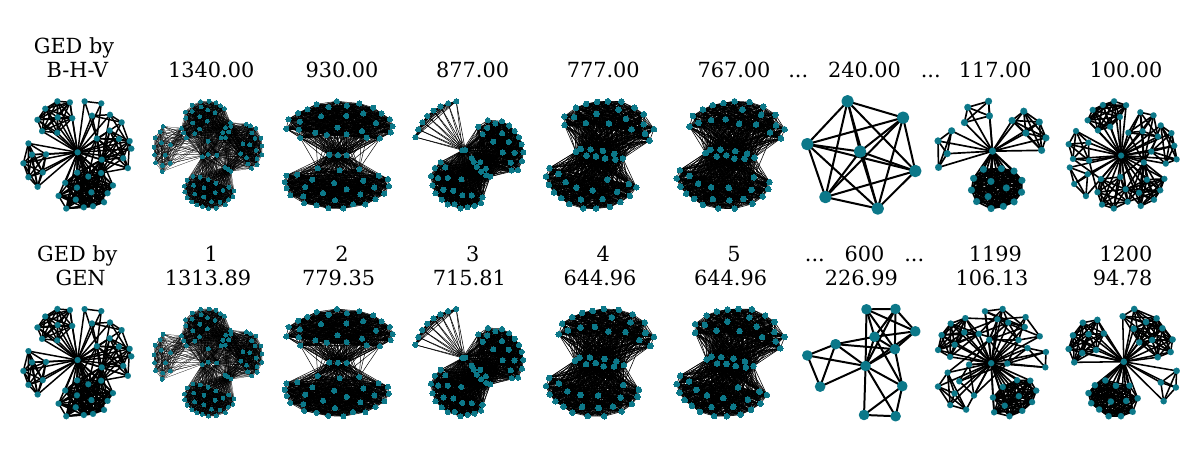}
    \caption{Case Study of ranking results on AIDS700nef (labeled) and IMDBMulti (unlabeled).}\label{fig:case}
\end{figure}
\begin{figure*}[t]
    \centering
    \begin{subfigure}[b]{0.329\textwidth}
        \centering
        \includegraphics[width=\textwidth]{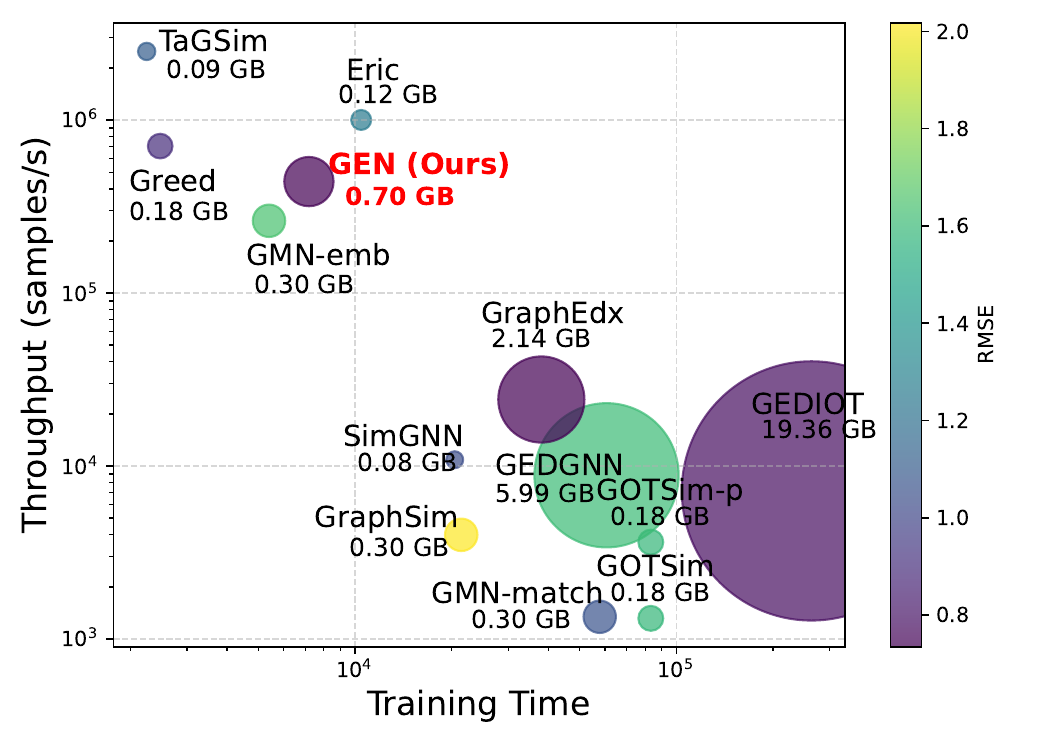}
        \caption{AIDS700nef}\label{fig:sub1}
    \end{subfigure}
    \begin{subfigure}[b]{0.329\textwidth}
        \centering
        \includegraphics[width=\textwidth]{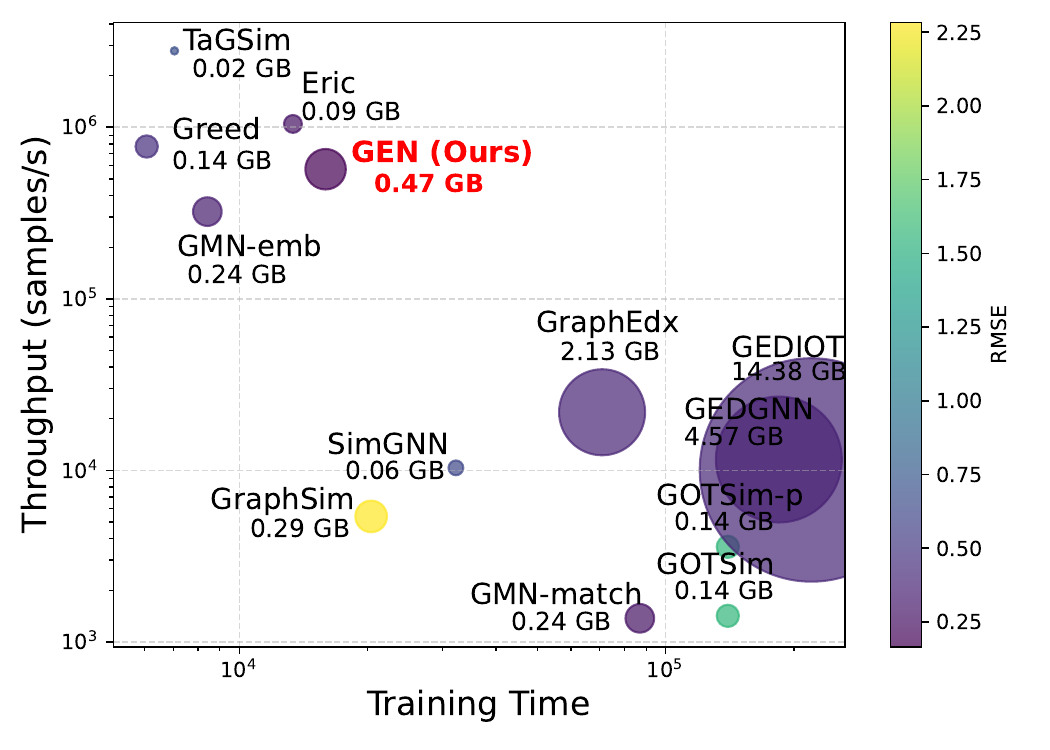}
        \caption{LINUX}
        \label{fig:sub2}
    \end{subfigure}
    \begin{subfigure}[b]{0.329\textwidth}
        \centering
        \includegraphics[width=\textwidth]{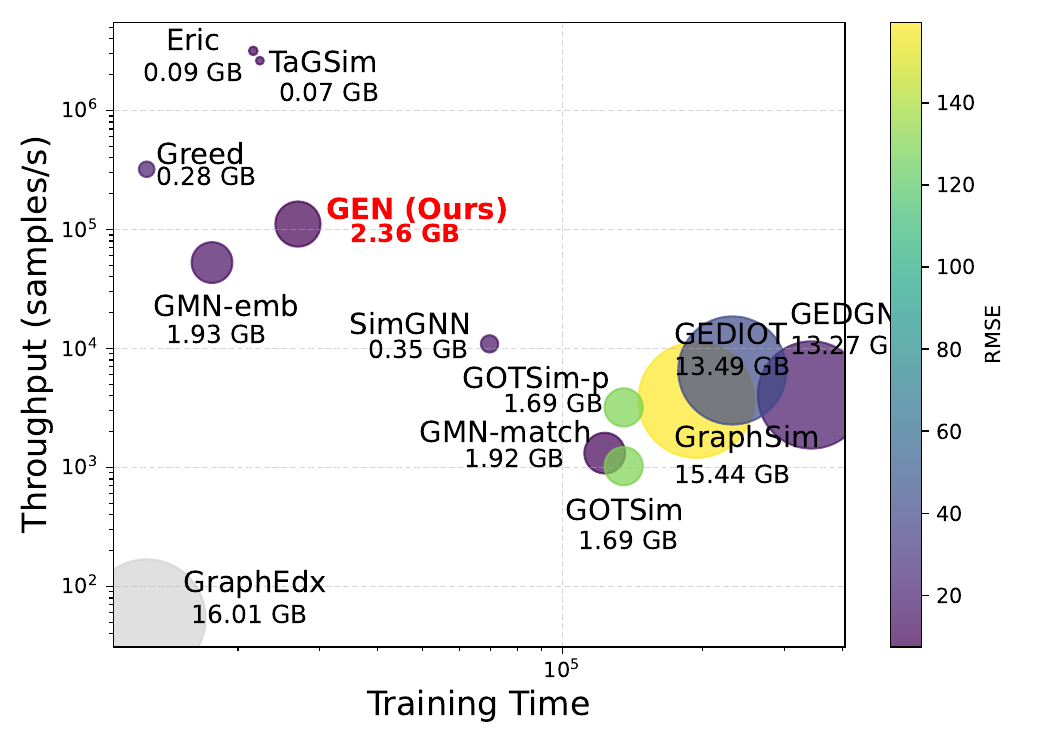}
        \caption{IMDBMulti}
        \label{fig:sub3}
    \end{subfigure}
    \caption{Training Time (x axis), Inference throughput (y axis) and Peak memory (bubble size).}
    \label{fig:efficiency}
\end{figure*}
\subsubsection{Case Study} 
In this experiment, we visualize the ranking results of queries on two datasets, AIDS700nef and IMDBMulti, which differ in both scale and structural characteristics. The objective is to evaluate the model’s ability to retrieve graphs with similar structures to the query based on their GED values within a graph search framework. As shown in Figure~\ref{fig:case}, \name\ exhibits a high degree of consistency with the rankings produced by the exact and inexact GED algorithms. This close alignment indicates that \name\ can effectively capture graph similarities and produce reliable retrieval results. To better understand the remaining ranking discrepancies, we further analyze the factors that may affect the ordering of retrieved graphs. We observe two main causes: 
\begin{itemize}
    \item Graph pairs with identical ground-truth GEDs may receive slightly different predicted scores due to continuous embeddings of one-hot node attributes, which the model can misinterpret as structural differences.
    \item Graph pairs containing cyclic structures may exhibit slight score deviations, reflecting the limited capacity of GNNs to fully capture such structures \cite{idgnn}.
\end{itemize}

\begin{table}[ht]
\centering
\caption{Model size and test batch size on IMDBMulti.}
\label{tab:method_scal}
\renewcommand{\arraystretch}{1.0}
\begin{tabular}{l|cc}
\toprule
\textbf{Method} & \textbf{Model size} &\textbf{batch size} \\
\midrule
\textbf{TaGSim} &
1,783 &
4,096 \\
\textbf{GMN-emb}  &
52,032 &
4,096 \\
\textbf{Greed} &
10,5698 &
4,096 \\
\textbf{Eric} &
84,309 &
4,096 \\
\midrule
\textbf{SimGNN} &
8,161 &
4,096 \\
\textbf{GMN-match} &
54,080 &
4,096 \\
\textbf{GraphSim} &
1,124,227 &
4,096 \\
\textbf{GOTSim} &
11,044 &
4,096 \\
\textbf{GEDGNN} &
86,614 &
512 \\
\textbf{GraphEdX} &
11,269 &
4 \\
\midrule
\textbf{Ours (\name)} &
63,569 &
4,096 \\
\bottomrule
\end{tabular}
\end{table}

\begin{table*}[!ht]
\caption{‘Rank' indicates the current ranking based on RMSE and show the improvements over the ranking for the AIDS700nef dataset in Table \ref{tab:aids_res} under a uniform cost setting, as shown in parentheses.}\label{tab:cost}
\resizebox{\textwidth}{!}{ \renewcommand{\arraystretch}{1.0}
\centering
\begin{tabular}{@{}l|cccc|c|cccc|c}
\toprule
\multirow{2}{*}{\textbf{Dataset}} &
\multicolumn{5}{c|}{\textbf{Setting 1}} &
\multicolumn{5}{c}{\textbf{Setting 2}} 
\\
\cmidrule(l{2pt}r{2pt}){2-11}
&
RMSE$\downarrow$ & MAE$\downarrow$ & $\rho$$\uparrow$ & $\tau$$\uparrow$ & Rank &
RMSE$\downarrow$ & MAE$\downarrow$ & $\rho$$\uparrow$ & $\tau$$\uparrow$ & Rank \\
\midrule
\textbf{SimGNN} &
$1.923_{\pm\text{0.082}}$ &
$1.487_{\pm\text{0.064}}$ &
$0.859_{\pm\text{0.010}}$ &
$0.680_{\pm\text{0.013}}$ &
5 (-1) &

$1.100_{\pm\text{0.024}}$ &
$0.862_{\pm\text{0.015}}$ &
$0.955_{\pm\text{0.002}}$ &
$0.820_{\pm\text{0.003}}$ &
4 (0) \\

\textbf{GMN-match} &
$4.874_{\pm\text{0.054}}$ &
$3.421_{\pm\text{0.054}}$ &
$0.423_{\pm\text{0.017}}$ &
$0.295_{\pm\text{0.014}}$ &
11 (-6)&

$4.761_{\pm\text{0.152}}$ &
$3.317_{\pm\text{0.086}}$ &
$0.471_{\pm\text{0.009}}$ &
$0.344_{\pm\text{0.009}}$ &
11 (-6)\\

\textbf{GMN-emb} &
$5.742_{\pm\text{0.309}}$ &
$4.070_{\pm\text{0.110}}$ &
$0.448_{\pm\text{0.009}}$ &
$0.311_{\pm\text{0.006}}$ &
13 (-3)&

$8.694_{\pm\text{0.117}}$ &
$6.093_{\pm\text{0.097}}$ &
$0.182_{\pm\text{0.028}}$ &
$0.125_{\pm\text{0.020}}$ &
13 (-3)\\

\textbf{GraphSim} &
$3.874_{\pm\text{1.636}}$ &
$2.707_{\pm\text{1.047}}$ &
$0.356_{\pm\text{0.419}}$ &
$0.283_{\pm\text{0.336}}$ &
9 (+2) &

$3.226_{\pm\text{1.574}}$ &
$2.488_{\pm\text{1.212}}$ &
$0.380_{\pm\text{0.458}}$ &
$0.320_{\pm\text{0.388}}$ &
9 (+2) \\

\textbf{GOTSim} &
$3.316_{\pm\text{0.211}}$ &
$2.635_{\pm\text{0.186}}$ &
$0.768_{\pm\text{0.058}}$ &
$0.577_{\pm\text{0.053}}$ &
8 (0) &

$2.754_{\pm\text{0.049}}$ &
$2.173_{\pm\text{0.034}}$ &
$0.760_{\pm\text{0.021}}$ &
$0.564_{\pm\text{0.017}}$ &
8 (0)\\

\textbf{TaGSim} &
$2.120_{\pm\text{0.051}}$ &
$1.605_{\pm\text{0.017}}$ &
$0.844_{\pm\text{0.003}}$ &
$0.662_{\pm\text{0.003}}$ &
6 (0) &

$1.188_{\pm\text{0.081}}$ &
$0.864_{\pm\text{0.032}}$ &
$0.950{\pm\text{0.004}}$ &
$0.816_{\pm\text{0.005}}$ &
6 (0) \\

\textbf{ERIC} &
$2.319_{\pm\text{0.173}}$ &
$1.826_{\pm\text{0.116}}$ &
$0.784_{\pm\text{0.035}}$ &
$0.599_{\pm\text{0.033}}$ &
7 (0) &

$1.587_{\pm\text{0.138}}$ &
$1.250_{\pm\text{0.096}}$ &
$0.911_{\pm\text{0.011}}$ &
$0.747_{\pm\text{0.014}}$ &
7 (0) \\

\textbf{Greed} &
$4.607_{\pm\text{0.045}}$ &
$3.396_{\pm\text{0.040}}$ &
$0.494_{\pm\text{0.012}}$ &
$0.348_{\pm\text{0.009}}$ &
10 (-7) &

$5.874_{\pm\text{0.072}}$ &
$4.200_{\pm\text{0.060}}$ &
$0.237_{\pm\text{0.013}}$ &
$0.170_{\pm\text{0.010}}$ &
12 (-9) \\

\textbf{GEDGNN} &
$1.782_{\pm\text{0.921}}$ &
$1.331_{\pm\text{0.598}}$ &
$0.916_{\pm\text{0.082}}$ &
$0.770_{\pm\text{0.104}}$ &
4 (+5) &

$1.102_{\pm\text{0.630}}$ &
$0.850_{\pm\text{0.466}}$ &
$0.953_{\pm\text{0.049}}$ &
$0.830_{\pm\text{0.080}}$ &
5 (+4) \\

\textbf{GraphEdX} &
$1.249_{\pm\text{0.014}}$ &
$0.978_{\pm\text{0.008}}$ &
$0.934_{\pm\text{0.002}}$ &
$0.786_{\pm\text{0.002}}$ &
3 (-2) &

$0.714_{\pm\text{0.012}}$ &
$0.567_{\pm\text{0.009}}$ &
$0.977_{\pm\text{0.001}}$ &
$0.871_{\pm\text{0.002}}$ &
3 (-2) \\

\midrule

\textbf{Ours (\name)} &
$\underline{1.061_{\pm\text{0.032}}}$ &
$\underline{0.771_{\pm\text{0.033}}}$ &
$\underline{0.952_{\pm\text{0.003}}}$ &
$\underline{0.823_{\pm\text{0.006}}}$ &
2 (0)&

$\underline{0.545_{\pm\text{0.007}}}$ &
$\underline{0.409_{\pm\text{0.007}}}$ &
$\underline{0.984_{\pm\text{0.000}}}$ &
$\underline{0.893_{\pm\text{0.001}}}$ &
2 (0) \\
\midrule
\midrule
\textbf{TaGSim mix} &
$7.106_{\pm\text{0.137}}$ &
$5.249_{\pm\text{0.159}}$ &
$0.307_{\pm\text{0.024}}$ &
$0.217_{\pm\text{0.017}}$ &
14 (new) &

$4.088_{\pm\text{0.171}}$ &
$3.087_{\pm\text{0.182}}$ &
$0.510_{\pm\text{0.014}}$ &
$0.382_{\pm\text{0.011}}$ &
10 (new) \\

\textbf{GraphEdX mix} &
$5.024_{\pm\text{0.411}}$ &
$4.081_{\pm\text{0.357}}$ &
$0.620_{\pm\text{0.058}}$ &
$0.446_{\pm\text{0.047}}$ &
12 (new) &

$12.918_{\pm\text{1.144}}$ &
$10.919_{\pm\text{1.334}}$ &
$0.218_{\pm\textbf{0.051}}$ &
$0.142_{\pm\textbf{0.038}}$ &
15 (new) \\

\midrule
\textbf{\name\ mix wo cost} &
${8.320}_{\pm\text{0.577}}$ &
${7.321}_{\pm\text{0.559}}$ &
${0.445}_{\pm\text{0.077}}$ &
${0.315}_{\pm\text{0.055}}$ &
15 (new) &

${7.415}_{\pm\text{0.677}}$ &
${6.585}_{\pm\text{0.601}}$ &
${0.328}_{\pm\text{0.127}}$ &
${0.205}_{\pm\text{0.114}}$ &
14 (new) \\
\midrule
\textbf{Ours (\name\space mix)} &
$\textbf{0.898}_{\pm\textbf{0.044}}$ &
$\textbf{0.642}_{\pm\textbf{0.043}}$ &
$\textbf{0.979}_{\pm\textbf{0.002}}$ &
$\textbf{0.881}_{\pm\textbf{0.006}}$ &
1 (new) &

$\textbf{0.479}_{\pm\textbf{0.013}}$ &
$\textbf{0.353}_{\pm\textbf{0.015}}$ &
$\textbf{0.988}_{\pm\textbf{0.001}}$ &
$\textbf{0.905}_{\pm\textbf{0.003}}$ &
1 (new) \\

\bottomrule
\end{tabular}
}

\end{table*}

\subsubsection{Time Efficiency} We evaluate the training time, inference throughput, and peak memory of all models. Results for VJ and Beam are excluded, as prior studies have shown that learning-based methods significantly outperform them in efficiency \cite{simgnn, h2mn}. Unlike conventional algorithms that require on-the-fly computation for each query, learning-based methods can generate indices for previously seen graphs, enabling efficient retrieval through reusable intermediate representations. This property makes them particularly suitable for high-volume small graph databases and high-frequency querying scenarios, where low latency and accurate ranking are prioritized over exact GED computation and edit paths.

Figure~\ref{fig:efficiency} shows the average total training time over five runs for all models with a training batch size of 64, reflecting both training speed and convergence efficiency. For GraphEdX, which encountered out-of-memory (OOM) errors on IMDBMulti during training, we assign its training time to that of the fastest model on this dataset for comparison. Inference throughput and peak memory are measured with a test batch size of 4,096. When OOM errors occur, the batch size is iteratively halved until successful execution. Model sizes and corresponding test batch sizes on IMDBMulti are summarized in Table~\ref{tab:method_scal}.

As shown in Figure~\ref{fig:efficiency}, our proposed \name\ achieves comparable training time and throughput to graph-level methods across all three datasets, surpassing the slowest graph-level method, GMN-emb, in throughput. Compared with node-level methods, \name\ trains substantially faster and infers more efficiently, highlighting the practical advantage of its one-step global-guided alignment. GOTSim, GEDGNN, and GraphEdX maintain relatively small parameter counts by trading off lightweight encoders for more sophisticated alignment techniques. GOTSim uses a CPU-based linear sum assignment solver, resulting in small memory usage. Its CPU multi-processing implementation (GOTSim-p) achieves an approximately $3\times$ throughput improvement. In contrast, GEDGNN and GraphEdX require the largest memory footprints due to iterative optimization with Gumbel-Sinkhorn networks.

Our \name\ represents an implementation of the revised paradigm. Compared with the most similar baseline, GMN-match, which has comparable overall performance, \name\ has up to 0.24× higher parameter count and 1.32× higher peak memory while delivering up to 414× higher inference throughput. Moreover, GMN-match iteratively updates and aligns node representations, producing pair-dependent embeddings that cannot be reused. In contrast, \name\ generates reusable, indexable node embeddings via a deeper neural encoder and performs an efficient one-step global-guided alignment, enabling high-throughput, low-latency inference suitable for high-volume or high-frequency graph retrieval scenarios.

\subsubsection{Handling Non-uniform and Mixed Costs} \label{ep:limit} 
We assess the capability of models to compute GED in a manner that respects the specified cost setting, focusing on non-uniform and mixed cost scenarios on the node-labeled AIDS700nef dataset. The purpose is to verify the effectiveness of incorporating operation costs into matching expense estimation, rather than to handle arbitrary unseen cost settings. Two cost settings are considered. In Setting 1, the costs for node substitution, node deletion, node insertion, edge deletion, and edge insertion are set to [1, 2, 1, 3, 1], respectively. In Setting 2, the corresponding costs are [3, 2, 3, 0, 2].
\begin{table}[!t]
\centering
\caption{Performance on an unseen cost setting.}
\label{tab:unseen}
\renewcommand{\arraystretch}{1.0}
\resizebox{\linewidth}{!}{
\begin{tabular}{@{}l|cccc@{}}
\toprule
\textbf{Model} & RMSE$\downarrow$ & MAE$\downarrow$ & $\rho$$\uparrow$ & $\tau$$\uparrow$ \\
\midrule
& \multicolumn{4}{c}{\textbf{Zero-shot Prediction}}\\
\midrule
\textbf{TaGSim mix}       &
$25.822_{\pm\text{0.658}}$ &
$22.558_{\pm\text{0.725}}$ &
$0.507_{\pm\text{0.030}}$ &
$0.372_{\pm\text{0.023}}$ \\
\textbf{GraphEdX mix}     &
$30.699_{\pm\text{1.770}}$ &
$27.793_{\pm\text{1.988}}$ &
$0.399_{\pm\text{0.049}}$ &
$0.280_{\pm\text{0.026}}$ \\
\textbf{Ours (\name\ mix)} &
$37.117_{\pm\text{0.675}}$ &
$34.605_{\pm\text{0.694}}$ &
$0.444_{\pm\text{0.044}}$ &
$0.312_{\pm\text{0.029}}$ \\
\midrule
& \multicolumn{4}{c}{\textbf{Few-shot Prediction}}\\
\midrule
\textbf{TaGSim}       &
$3.816_{\pm\text{0.177}}$ &
$1.563_{\pm\text{0.074}}$ &
$0.967_{\pm\text{0.002}}$ &
$0.894_{\pm\text{0.003}}$ \\
\textbf{GraphEdX}     &
$6.619_{\pm\text{0.956}}$ &
$4.628_{\pm\text{0.786}}$ &
$0.920_{\pm\text{0.032}}$ &
$0.768_{\pm\text{0.043}}$ \\
\textbf{Ours (\name)} &
$1.414_{\pm\text{0.027}}$ &
$0.734_{\pm\text{0.015}}$ &
$0.995_{\pm\text{0.000}}$ &
$0.945_{\pm\text{0.001}}$ \\
\midrule
& \multicolumn{4}{c}{\textbf{Few-shot Finetune}}\\
\midrule
\textbf{Ours (\name\ mix)} &
$1.584_{\pm\text{0.028}}$ &
$0.971_{\pm\text{0.025}}$ &
$0.994_{\pm\text{0.000}}$ &
$0.938_{\pm\text{0.001}}$ \\
\bottomrule
\end{tabular}}
\vspace{-0.5cm}
\end{table}

\begin{table*}[!ht]
\centering
\caption{Ablations Study on our proposed components.}
\resizebox{0.75\linewidth}{!}{ \renewcommand{\arraystretch}{1.0}
\centering
\begin{tabular}{@{}l|cc|cc|cc}
\toprule
\multirow{2}{*}{\textbf{Dataset}} &
\multicolumn{2}{c}{\textbf{AIDS700nef}} &
\multicolumn{2}{c}{\textbf{LINUX}} &
\multicolumn{2}{c}{\textbf{IMDBMulti}} \\
&
RMSE $\downarrow$ &
p@10 $\uparrow$ &
RMSE $\downarrow$ &
p@10 $\uparrow$ &
RMSE $\downarrow$ &
p@10 $\uparrow$ \\
\midrule
\textbf{\name\space wo global} &
$0.808_{\pm\text{0.038}}$ &
$0.954_{\pm\text{0.009}}$ &

$0.178_{\pm\text{0.013}}$ &
$0.993_{\pm\text{0.005}}$ &

$9.415_{\pm\text{0.977}}$ &
$0.972_{\pm\text{0.023}}$ \\
\textbf{\name\space wo dependencies} &
$1.101_{\pm\text{0.041}}$ &
$0.905_{\pm\text{0.006}}$ &

$0.296_{\pm\text{0.034}}$ &
$\underline{0.990_{\pm\text{0.005}}}$ &

$14.391_{\pm\text{2.568}}$ &
$0.982_{\pm\text{0.004}}$ \\
\textbf{\name\space wo intra-graph} &
$\underline{0.746_{\pm\text{0.011}}}$ &
$\underline{0.958_{\pm\text{0.002}}}$ &

$\textbf{0.159}_{\pm\textbf{0.015}}$ &
$\textbf{0.995}_{\pm\textbf{0.003}}$ &

$\underline{8.439_{\pm\text{0.656}}}$ &
$\underline{0.984_{\pm\text{0.005}}}$ \\
\midrule
\textbf{\name} &
$\textbf{0.738}_{\pm\textbf{0.007}}$ &
$\textbf{0.963}_{\pm\textbf{0.003}}$ &

$\underline{0.166_{\pm\text{0.009}}}$ &
$\textbf{0.995}_{\pm\textbf{0.004}}$ &

$\textbf{7.531}_{\pm\textbf{0.293}}$ &
$\textbf{0.987}_{\pm\textbf{0.001}}$ \\
\bottomrule
\end{tabular}
}
\label{tab:ab}
\end{table*}

As shown in Table~\ref{tab:cost}, \textbf{(i) under non-uniform costs}, \name\ consistently achieves the best performance in both settings. This advantage stems from explicitly incorporating operation costs into the node matching expense estimation, ensuring that the learned alignments faithfully reflect the specified cost scheme. In contrast, Greed performs poorly due to its reliance on a symmetric distance measure, while GMN-match suffers from accumulated errors caused by its greedy local fusion mechanism. Notably, node-level models demonstrate more significant rank improvements than graph-level ones, indicating their superior suitability for cost-aware GED estimation. 

\textbf{(ii) Under mixed costs}, we train \name\ mix, \name\ mix wo cost (without cost awareness), TaGSim mix, and GraphEdX mix on a dataset combining both cost settings. As shown in Table~\ref{tab:cost}, TaGSim mix and GraphEdX mix exhibit sharp performance degradation because they ignore operation costs during matching expense estimation, resulting in node alignments that are insensitive to the cost setting. In contrast, \name\ mix maintains stable performance and even improves over \name, whereas \name\ mix without cost awareness performs substantially worse, demonstrating that explicitly integrating operation costs ensures the resulting alignment reflects the intended cost setting. This highlights the fundamental limitation of existing paradigms that treat cost as an external post-processing factor. 

\textbf{(iii) Adapt to unseen costs}, we evaluate \name\ mix, TaGSim mix, and GraphEdX mix under an unseen cost setting where the costs are [4, 1, 2, 2, 3]. As shown in Table~\ref{tab:unseen}, all methods perform poorly in zero-shot evaluation, reflecting a general challenge in GED estimation under varying cost settings. TaGSim and GraphEdX achieving relatively better zero-shot performance due to their cost-independent matching mechanisms. To further investigate, we perform few-shot training using 10\% of the data under the new cost setting for \name, TaGSim, and GraphEdX. Additionally, we finetune \name\ mix by updating only the MLP cost estimator in Equation (\ref{eq:estimator}), adjusting cost-aware matching expenses without altering previously learned representations. After adaptation, few-shot trained \name\ achieves the best performance, while finetuned \name\ mix performs comparably and surpasses the few-shot trained baselines. These results demonstrate that \name\ can efficiently adapt to new cost settings with minimal intervention, a particularly valuable property given the computational cost of generating ground-truth GED for unseen costs.

These results validate the effectiveness of our revision, demonstrate the necessity of integrating operation costs into matching expense estimation, and suggest a future direction: designing architectures that can dynamically adjust matching expenses for unseen cost settings.

\subsubsection{Ablation Study} We evaluate the effectiveness of our proposed global guidance. We test three variants: 
\begin{enumerate}
    \item \name\space wo global: This variant replaces the global cost ratio with absolute node matching costs to estimate alignment difficulties.
    \item \name\space wo dependencies: This variant reweighs matches solely based on their global cost ratios, where higher ratios result in lower weights.
    \item \name\space wo intra-graph: This variant does not consider intra-graph connections.
\end{enumerate}

Note that the inter-graph discrepancies cannot be ablated separately as they produce the matching decisions. 
As shown in Table \ref{tab:ab}, \name\space wo dependencies greedily reweighs matches based on global cost ratios, leading to a significant performance drop across all datasets, with RMSE increasing by up to 91.1\%. On the other hand, \name\space wo global incurs a smaller increase, at most 25\%, indicating that considering dependencies is more crucial than the global cost ratio. Meanwhile, ignoring the difficulties of neighboring nodes has a smaller impact. \name\space wo intra-graph showcased comparable performance on AIDS700nef and LINUX, yet incurred a 12.1\% increase in RMSE on IMDBMulti. This is likely because the matches in IMDBMulti are more ambiguous, making the neighborhood alignment difficulty a crucial factor.

\begin{figure}[]
    \begin{subfigure}{\linewidth}
    \centering
    \includegraphics[width=\linewidth]{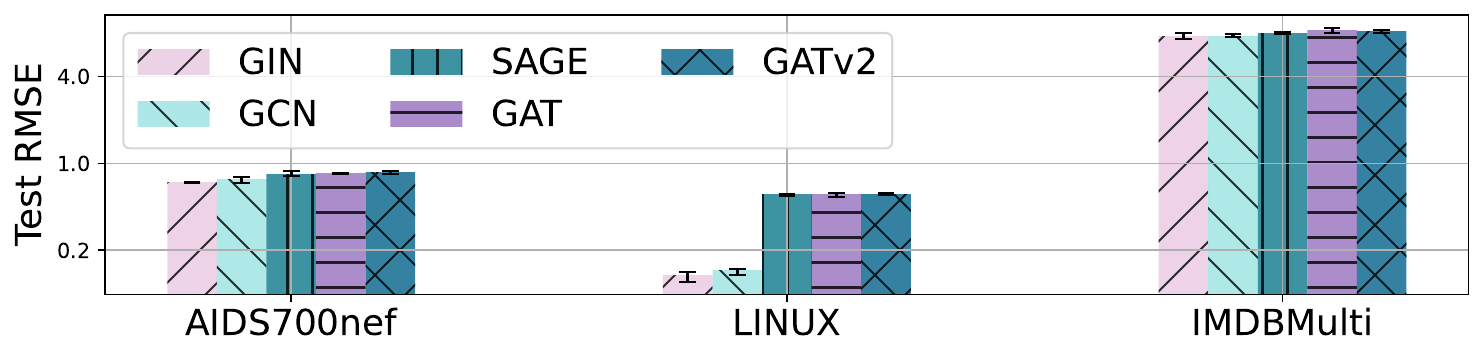}
    \end{subfigure}

    \begin{subfigure}{0.49\linewidth}
    \centering
    \includegraphics[width=\linewidth]{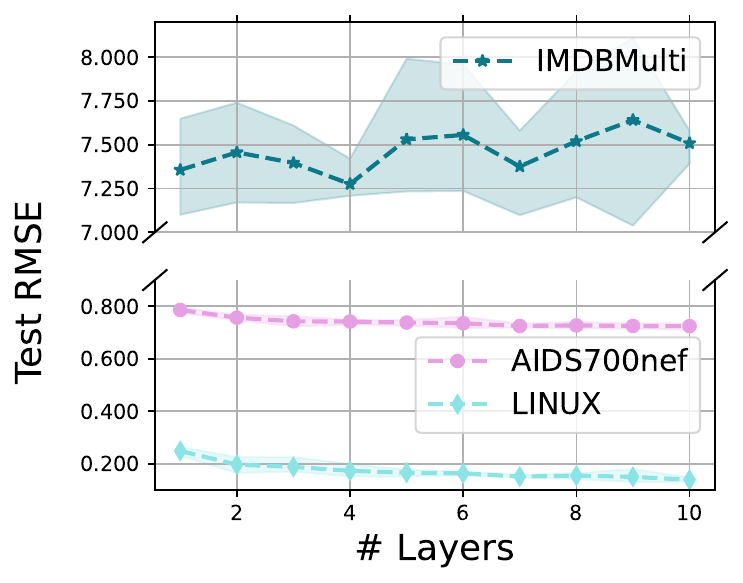}
    \end{subfigure}
    \begin{subfigure}{0.49\linewidth}
    \centering
    \includegraphics[width=\linewidth]{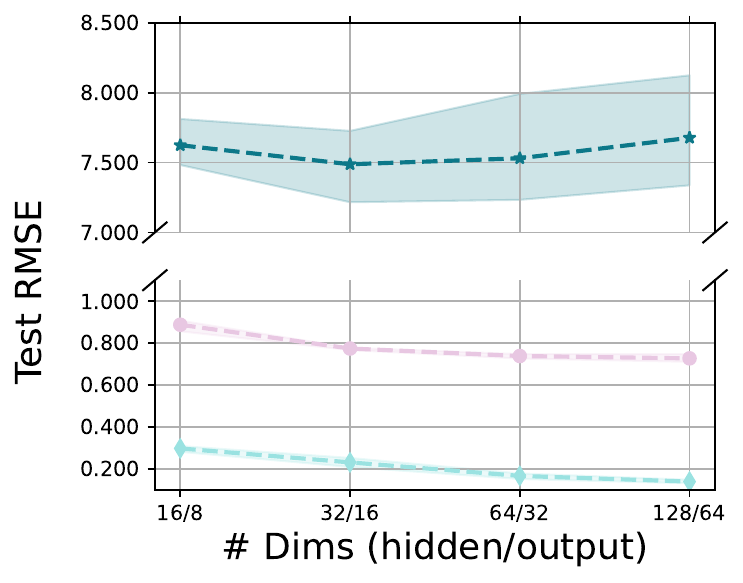}
    \end{subfigure}
    \caption{Hyperparameter sensitivity.}\label{fig:t_layers}
    \vspace{-0.5cm}
\end{figure}

\begin{figure*}[t]
    \centering
    \begin{subfigure}[b]{0.329\textwidth}
        \centering
        \includegraphics[width=\textwidth]{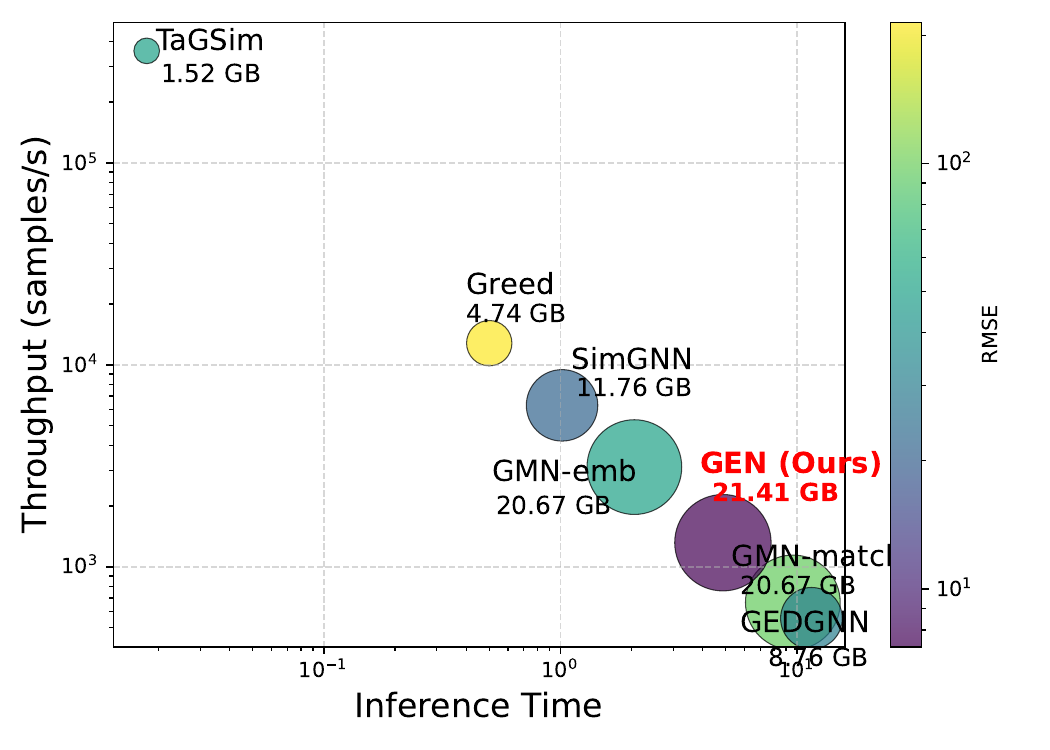}
        \caption{[100, 200]}\label{fig:sub1}
    \end{subfigure}
    \begin{subfigure}[b]{0.329\textwidth}
        \centering
        \includegraphics[width=\textwidth]{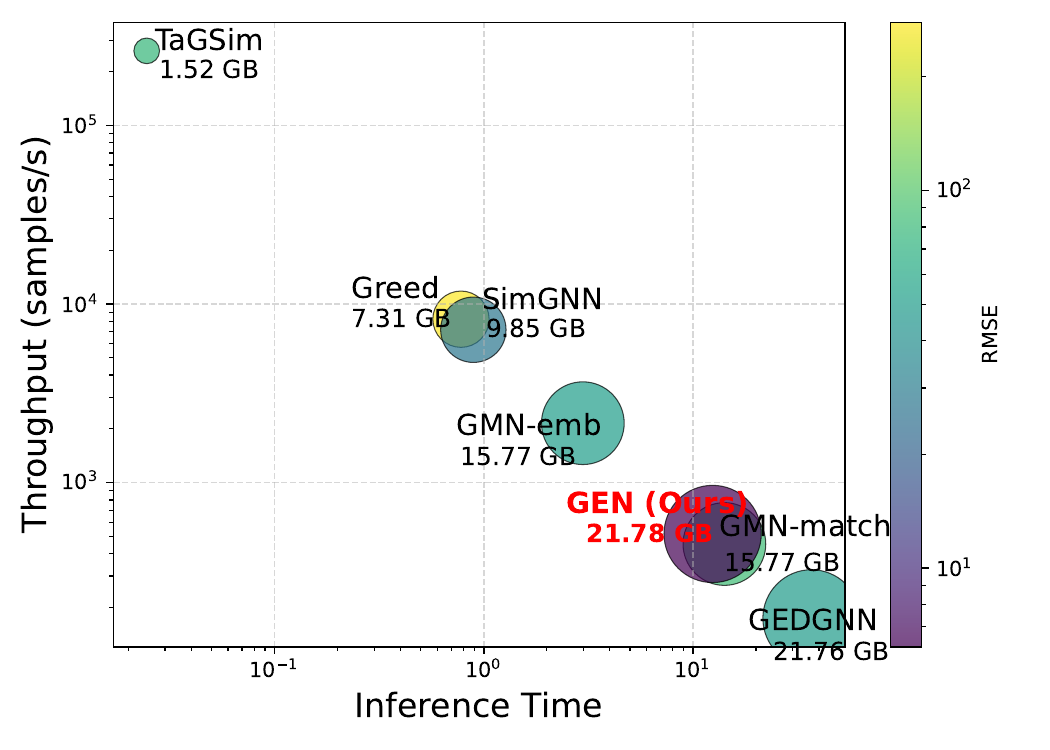}
        \caption{[200, 300]}
        \label{fig:sub2}
    \end{subfigure}
    \begin{subfigure}[b]{0.329\textwidth}
        \centering
        \includegraphics[width=\textwidth]{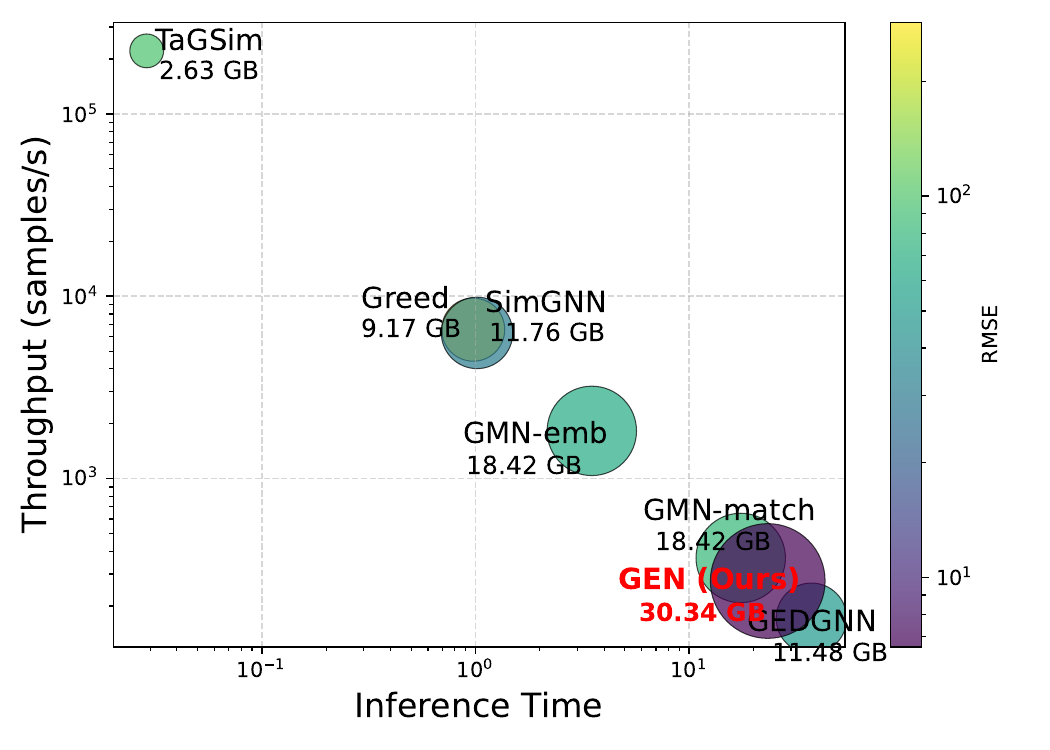}
        \caption{[300, 400]}
        \label{fig:sub3}
    \end{subfigure}
    \caption{Inference Time (x axis), Inference throughput (y axis) and Peak memory (bubble size).}
    \label{fig:efficiency_scal}
\end{figure*}
\begin{figure*}[!htb]
    \centering
    \begin{subfigure}{0.16\linewidth}
    \centering
    \includegraphics[width=\linewidth]{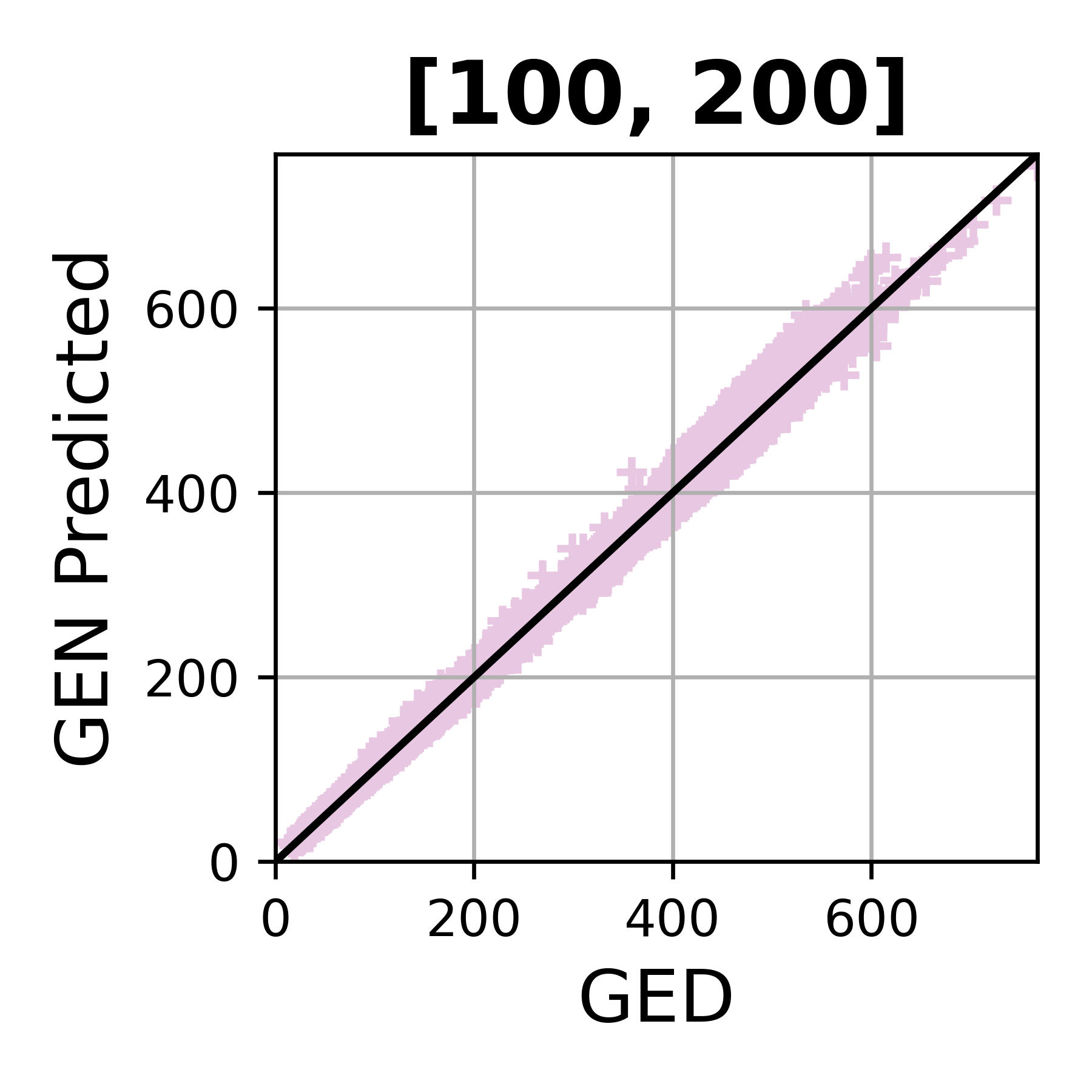}
    \end{subfigure}
    \begin{subfigure}{0.16\linewidth}
    \centering
    \includegraphics[width=\linewidth]{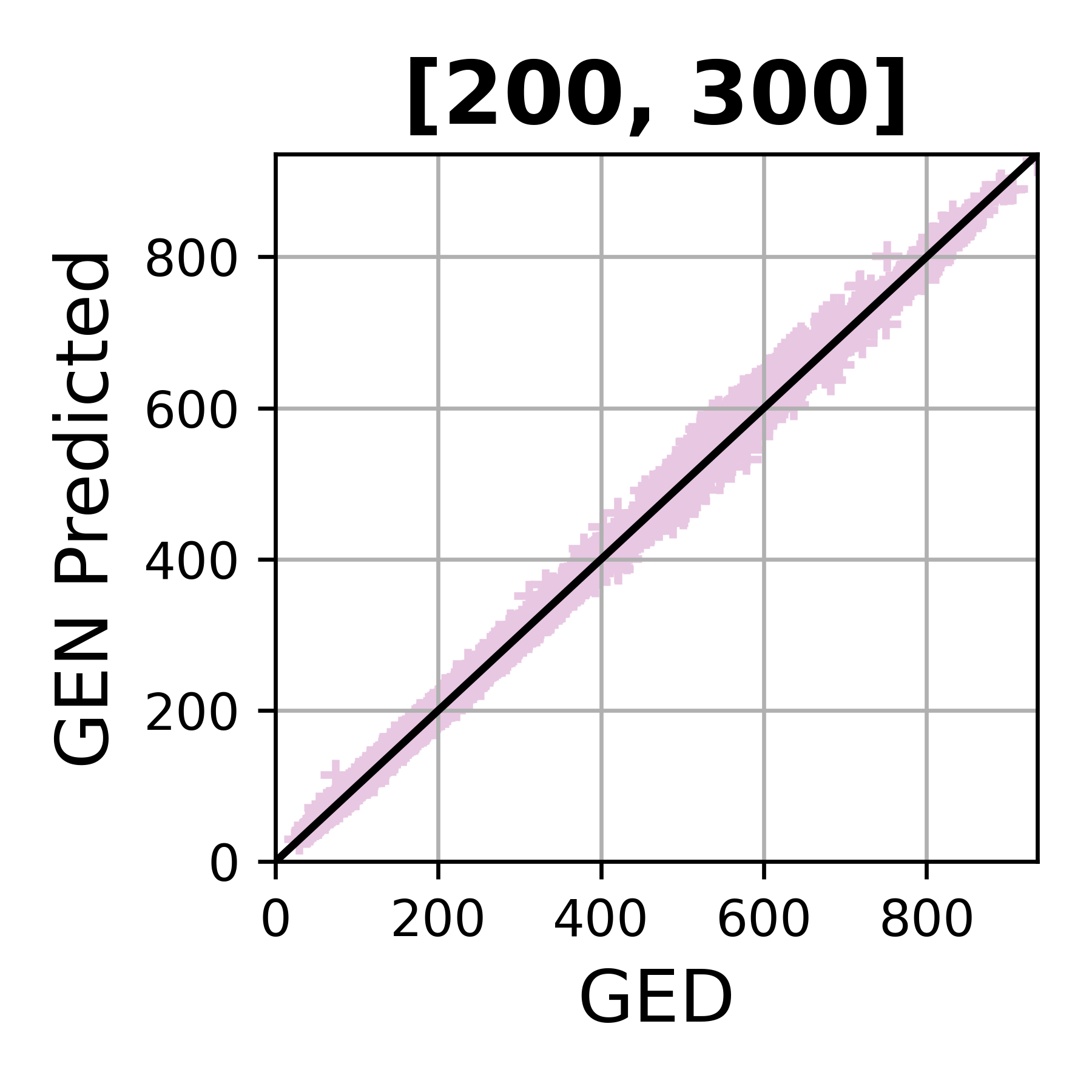}
    \end{subfigure}
    \begin{subfigure}{0.16\linewidth}
    \centering
    \includegraphics[width=\linewidth]{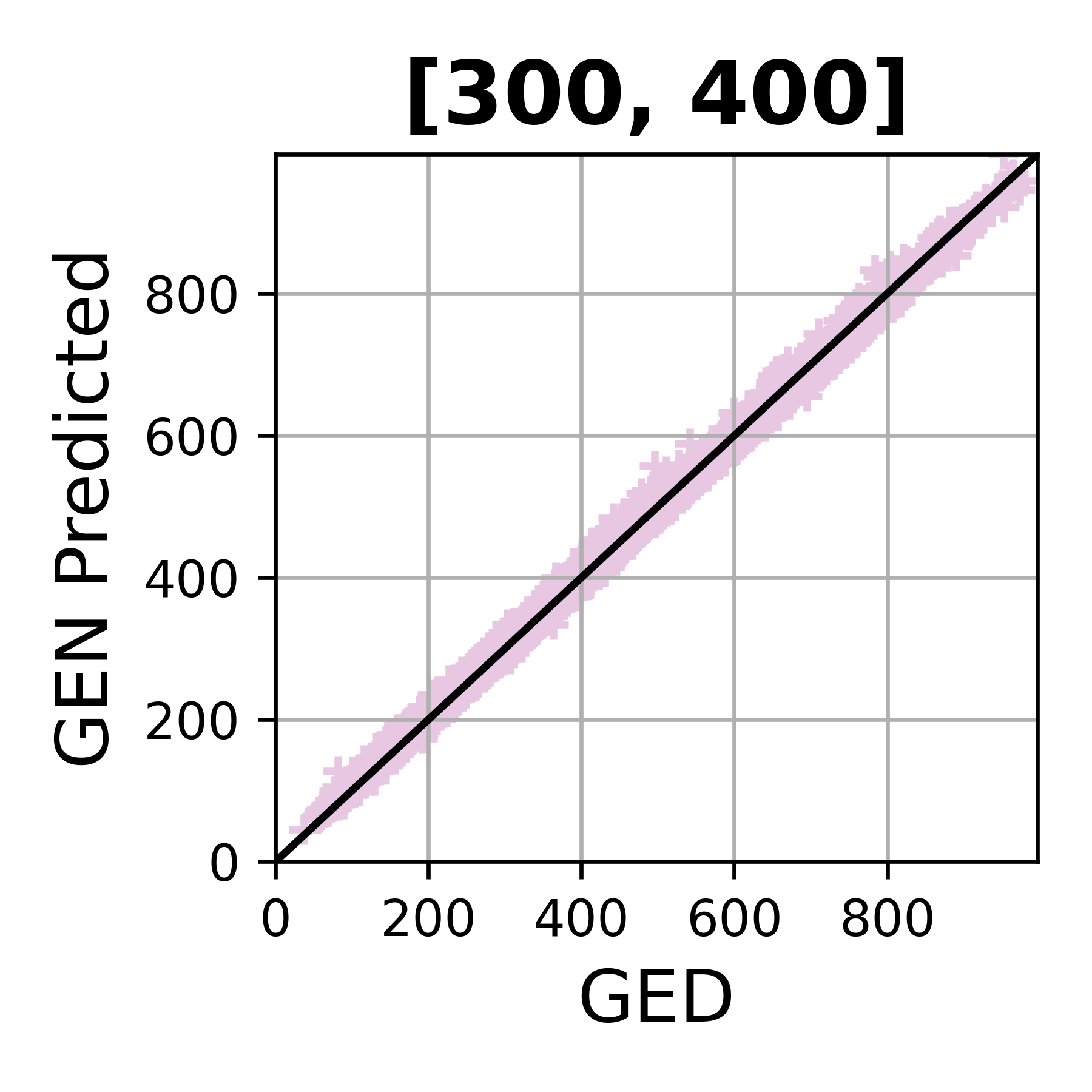}
    \end{subfigure}
    \begin{subfigure}{0.16\linewidth}
    \centering
    \includegraphics[width=\linewidth]{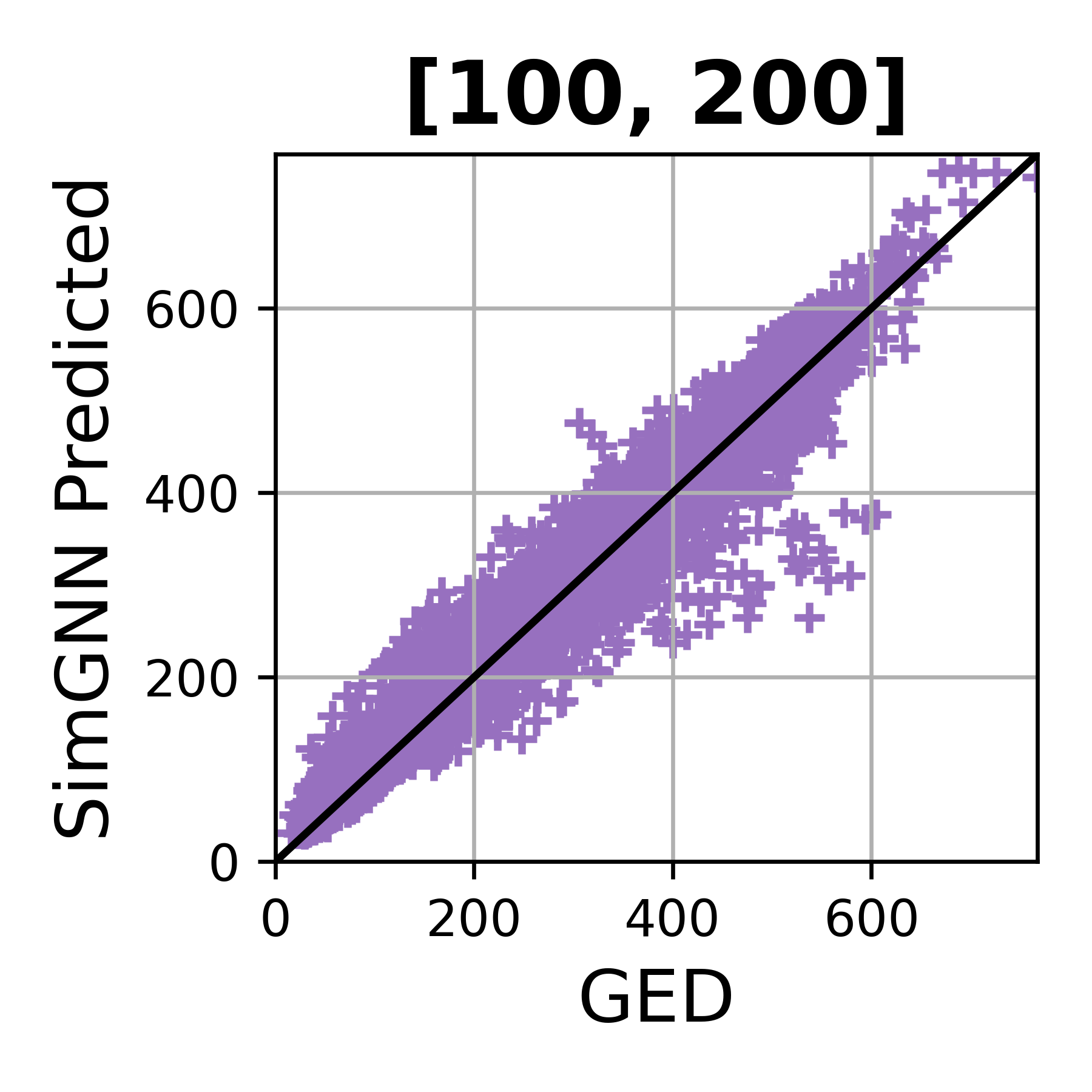}
    \end{subfigure}
    \begin{subfigure}{0.16\linewidth}
    \centering
    \includegraphics[width=\linewidth]{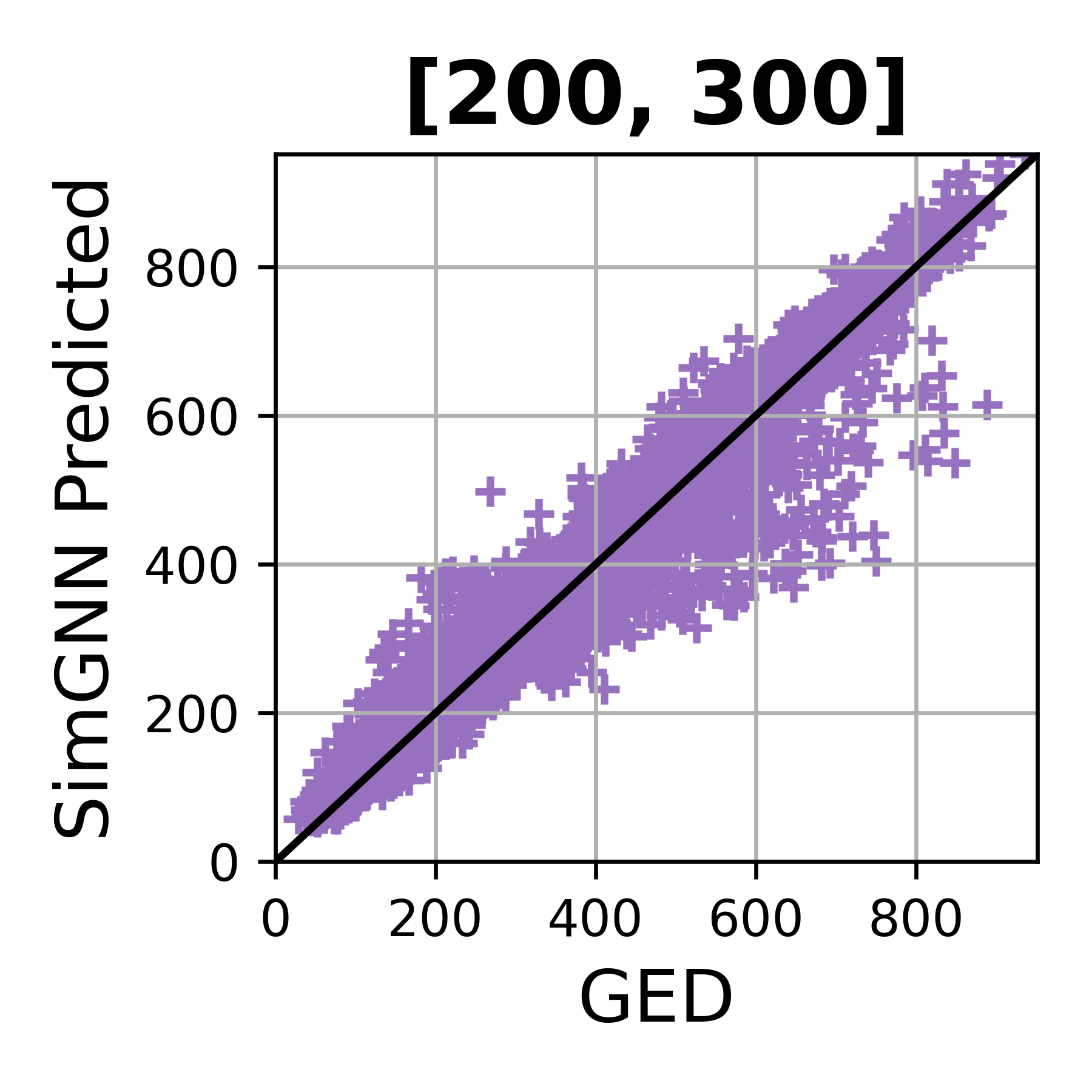}
    \end{subfigure}
    \begin{subfigure}{0.16\linewidth}
    \centering
    \includegraphics[width=\linewidth]{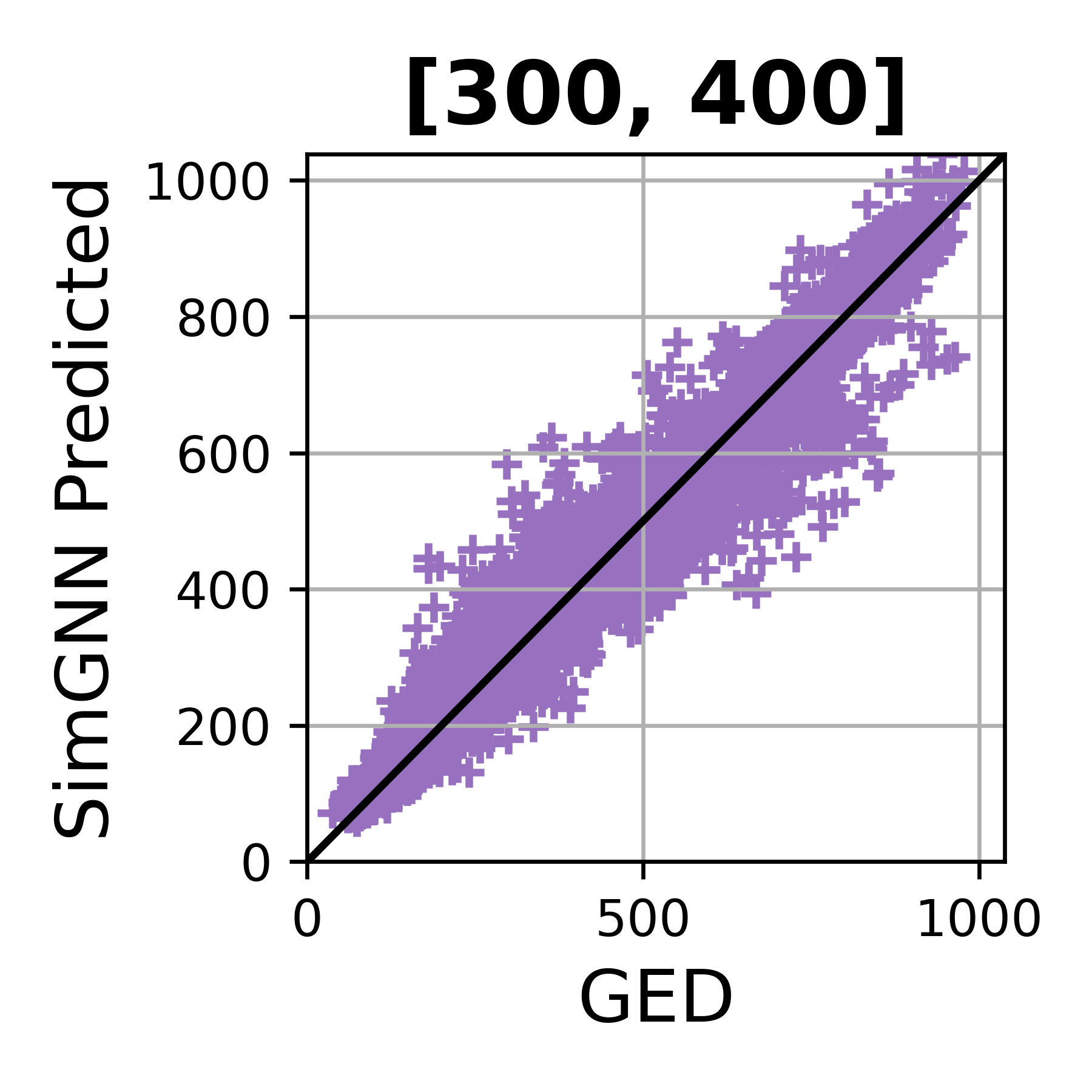}
    \end{subfigure}
    \caption{Performance of models on larger synthetic graphs across diverse node size groups.}\label{fig:errors}
\end{figure*}
\subsubsection{Hyperparameter Sensitivity} We analyze \name's sensitivity to key hyperparameters such as GNN backbones, number of layers, and dimensionality. As shown in Figure \ref{fig:t_layers}, GIN proves to be the optimal backbone for \name, consistently achieving the lowest RMSE. Adding layers generally improves performance, with the effect varying across datasets. On AIDS700nef and LINUX, performance increases steadily with more layers, while on IMDBMulti, it peaks at four layers for RMSE reduction. Similarly, increasing dimensionality typically reduces RMSE, but on IMDBMulti, performance worsens beyond 32/16 dimensions, likely due to overfitting.

\subsubsection{Performance on Larger Synthetic Graphs} GED is typically applied to small graphs, as exact computation becomes intractable for graphs with more than 16 nodes \cite{exact}. Following \cite{gedgnn}, we evaluate performance on larger synthetic graphs. Since synthetic graphs cannot fully capture real-world structural heterogeneity, in this experiment we assess predictive accuracy, inference time, throughput, and peak memory under extreme conditions to provide an estimate of expected behavior on larger graphs.The results in Figure~\ref{fig:efficiency_scal} show that GEN maintains stable accuracy as graph size increases, while most baselines exhibit growing prediction errors. Compared with node-level methods such as SimGNN and GMN-match, the introduction of global guidance in GEN increases memory usage, which necessitates smaller test batch sizes as graph size grows and consequently reduces throughput. To further illustrate the predictive advantage of GEN, we compare its predictions with those of the next-best model, SimGNN, against the ground-truth GEDs. As shown in Figure~\ref{fig:errors}, GEN’s predictions are tightly clustered around the line representing perfect alignment with ground truth across all node-size groups, whereas SimGNN’s predictions are more scattered.
\section{Conclusion}
This paper highlights three key factors in candidate matching for Graph Edit Distance (GED) computation: intrinsic differences, operation costs, and matching dependencies. It shows that existing learning-based fine-grained methods follow the prevailing paradigm, decoupling local matching selections from operation costs and match dependencies, which significantly undermines their capacity to make informed matching decisions. As a result, these methods often produce alignments that are independent of the cost settings and require costly iterative refinement to improve performance. To address this, we presents a revised formulation of the prevailing paradigm and propose Graph Edit Network (\name), which integrates cost-aware matching expense estimation with globally guided one-step alignment, enabling accurate GED approximation under specified cost settings without iterative refinements. Experiments on real and synthetic benchmarks demonstrate substantial improvements in accuracy and inference throughput, highlighting \name’s practicality for high-volume graph collections and high-frequency querying. Beyond this implementation, the proposed revision provides a principled perspective that can inform the design of more general and adaptable learning-based GED approximation frameworks.
\section*{AI-Generated Content Acknowledgment}
The authors acknowledge the use of large language model to improve the language presentation of this manuscript. The AI tool was not used for generating or validating any scientific results, analyses, or claims.
\bibliographystyle{IEEEtran}
\bibliography{references}

\end{document}